\documentclass[letterpaper, 10 pt, conference]{ieeeconf}  %

\IEEEoverridecommandlockouts                              %

\usepackage{times}
\usepackage{epsfig}
\usepackage{graphicx}
\usepackage{amsmath}
\usepackage{amssymb}

\usepackage{subfigure}
\usepackage{lipsum}

\title{\LARGE \bf
Fast Panoptic Segmentation Network
}

\author{Daan de Geus, Panagiotis Meletis, and Gijs Dubbelman%
\thanks{Daan de Geus, Panagiotis Meletis, and Gijs Dubbelman are with the Mobile Perception Systems research lab of the SPS/VCA group, Department of Electrical Engineering, Eindhoven University of Technology, The Netherlands
        {\{\tt\small d.c.d.geus}, {\tt\small p.c.meletis}, {\tt\small g.dubbelman}\}{\tt\small @tue.nl}}%
}

\begin{document}

\maketitle
\thispagestyle{empty}
\pagestyle{empty}

\begin{abstract}
In this work, we present an end-to-end network for fast panoptic segmentation. This network, called Fast Panoptic Segmentation Network (FPSNet), does not require computationally costly instance mask predictions or merging heuristics. This is achieved by casting the panoptic task into a custom dense pixel-wise \textit{classification} task, which assigns a class label or an instance \textit{id} to each pixel. We evaluate FPSNet on the Cityscapes and Pascal VOC datasets, and find that FPSNet is faster than existing panoptic segmentation methods, while achieving better or similar panoptic segmentation performance. On the Cityscapes validation set, we achieve a Panoptic Quality score of 55.1\%, at prediction times of 114 milliseconds for images with a resolution of 1024x2048 pixels. For lower resolutions of the Cityscapes dataset and for the Pascal VOC dataset, FPSNet runs at 22 and 35 frames per second, respectively.

\end{abstract}

\section{Introduction}

Panoptic segmentation \cite{Kirillov2018} is a task for which the goal is to predict a class label and an instance \textit{id} for each pixel in an image. A distinction is made between \textit{things} and \textit{stuff} classes. For \textit{things} classes, which have countable objects (e.g. person, car), the instance \textit{id} is used to distinguish between different objects, whereas all \textit{stuff} classes receive the same instance \textit{id}, as these parts of the image are usually uncountable (e.g. sky, water). In this work, we present an end-to-end deep neural network architecture for fast panoptic segmentation, that is able to achieve real-time inference speeds.

Panoptic segmentation is closely related to the tasks of semantic segmentation and instance segmentation. For semantic segmentation, the goal is to predict a class label -- for both \textit{stuff} and \textit{things} classes -- for each pixel in an image, whereas instance segmentation aims at finding pixel-level masks for all \textit{things} instances in an image. Current panoptic segmentation methods~\cite{Kirillov2019, Xiong2019, Porzi2019Seamless, Li2018TASCNet, Li2018AUNet} exploit this relation between these tasks. Instead of training the panoptic task directly, i.e. fully end-to-end, they train the instance segmentation and semantic segmentation tasks separately, and fuse the outputs into the panoptic format. This requires solving conflicts between instance segmentation and semantic segmentation predictions. Firstly, instance segmentation predictions can overlap each other, and secondly, pixels can also get different predictions from the instance segmentation and semantic segmentation output. These conflicts are problematic because panoptic segmentation allows for only one prediction per pixel. Current state-of-the-art methods rely on heuristics to resolve these conflicts \cite{Kirillov2019, Porzi2019Seamless, Li2018AUNet}. We propose a fast, heuristic-free approach that is able to learn to resolve these conflicts.

\begin{figure}[t]
\centering
\includegraphics[width=1.0\linewidth]{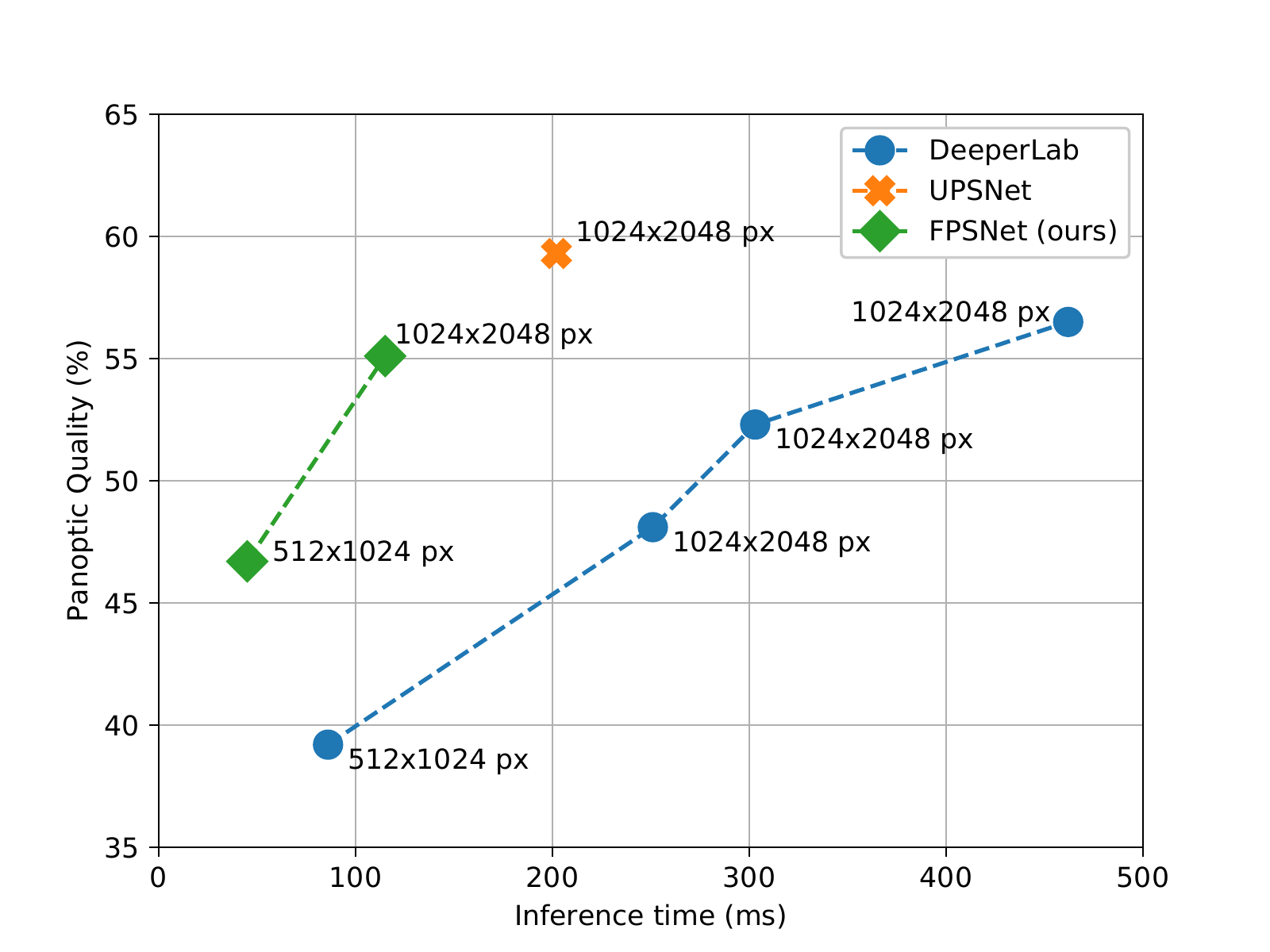}
\caption{Prediction time vs. Panoptic Quality for various methods on the Cityscapes validation set. We also indicate the input resolution.}
\label{fig:time_pq_plot}

\end{figure}

\begin{figure}[t]
\centering
\subfigure[Input image]{\includegraphics[width=0.490\linewidth]{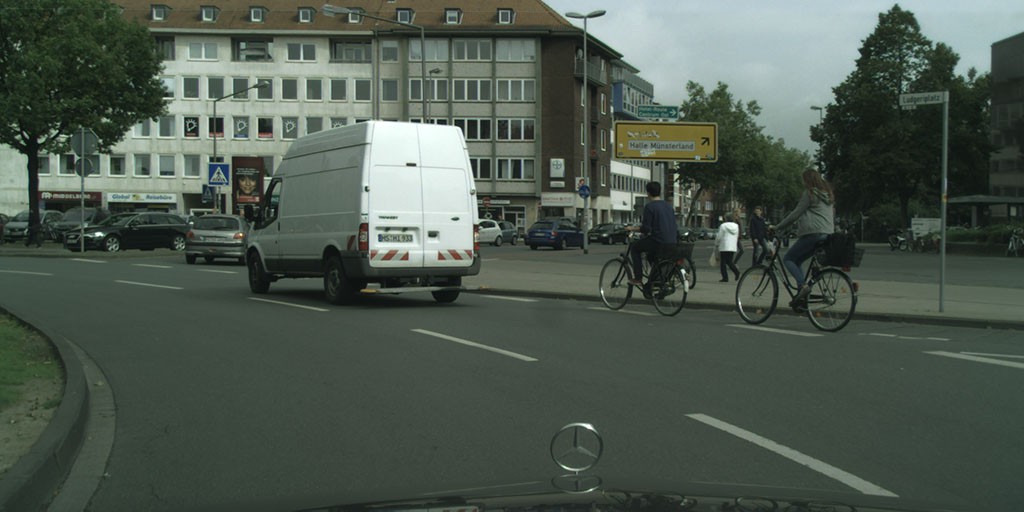}}
\subfigure[Ground truth]{\includegraphics[width=0.490\linewidth]{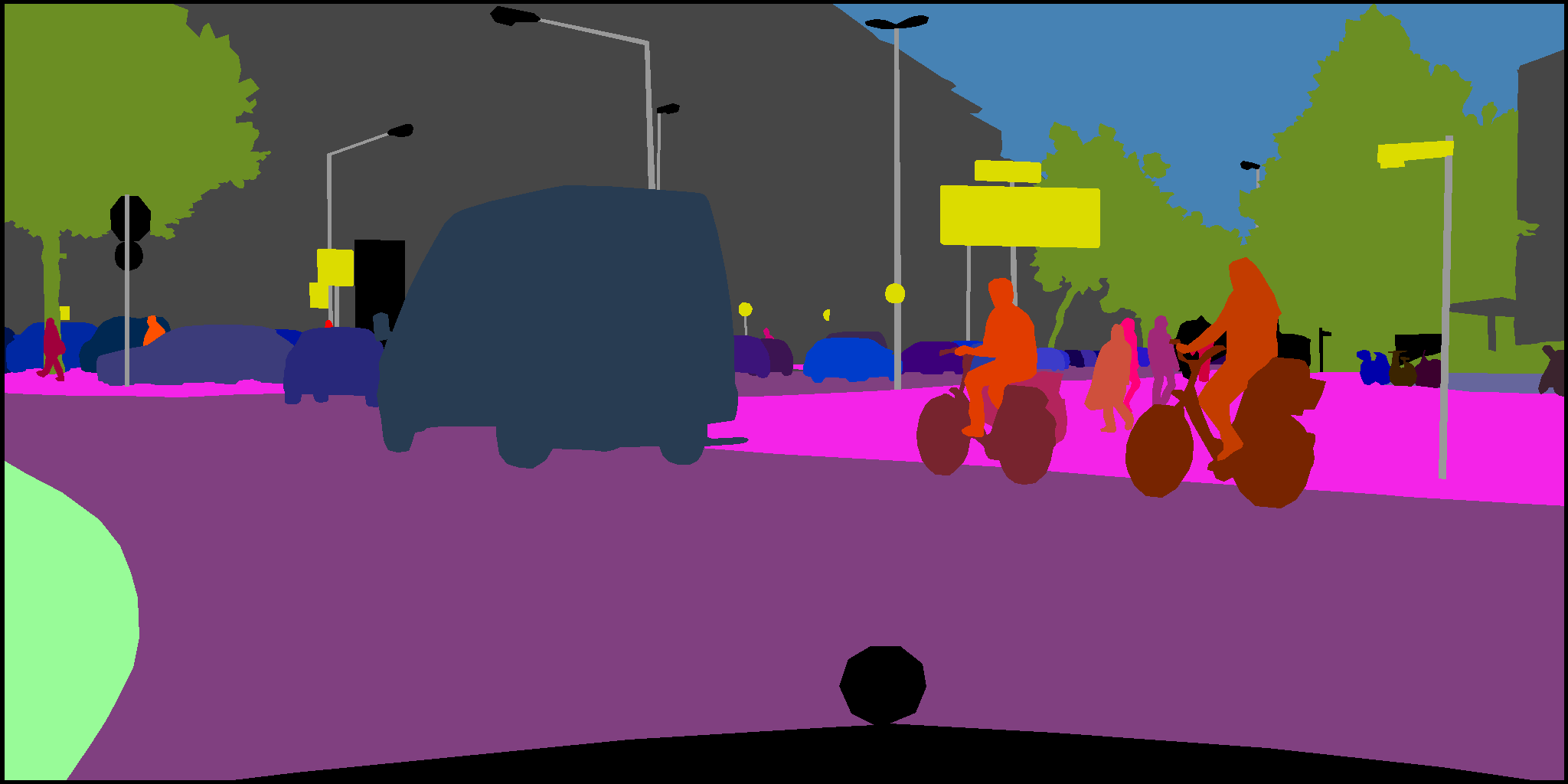}}\\
\subfigure[Inference at 1024x2048 px]{\includegraphics[width=0.490\linewidth]{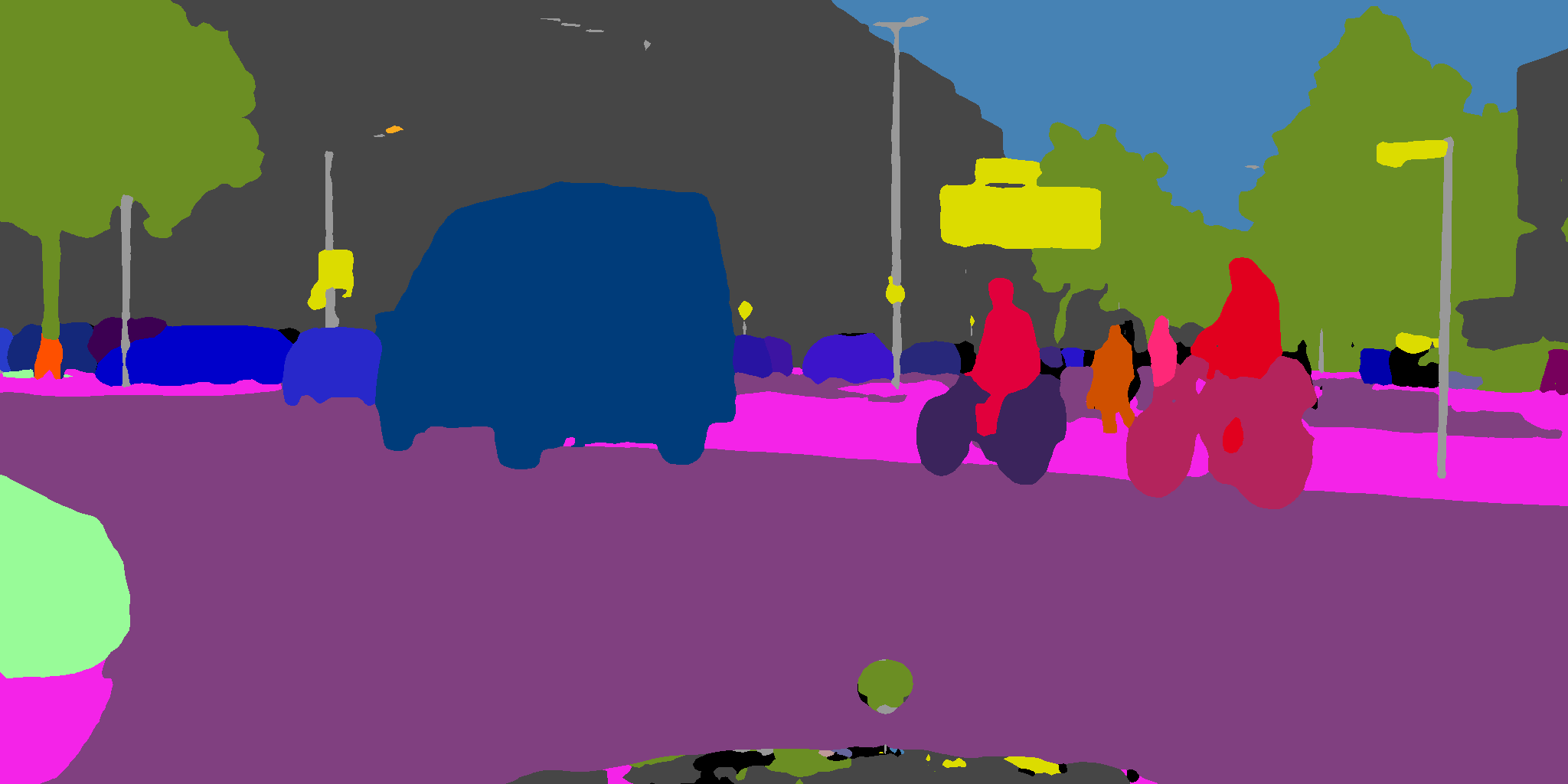}}
\subfigure[Inference at 512x1024 px ]{\includegraphics[width=0.490\linewidth]{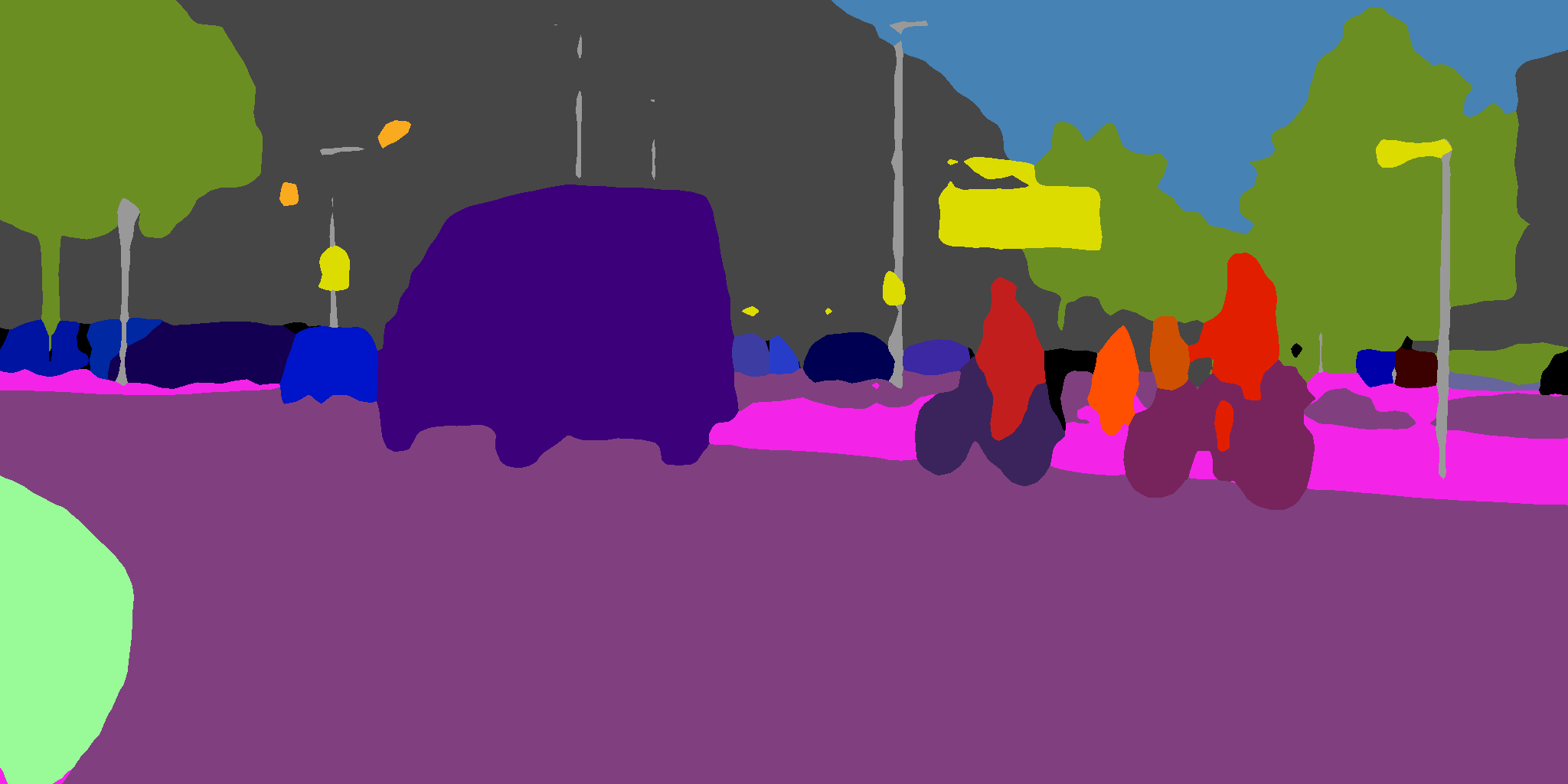}}
\caption{FPSNet predictions on image from the Cityscapes validation set at different input resolutions. Each color indicates a different \textit{things} instance or \textit{stuff} class.} 
\label{fig:main_pred}
\vspace{-10px}

\end{figure}

\begin{figure*}[t]
\includegraphics[width=1.0\linewidth, clip]{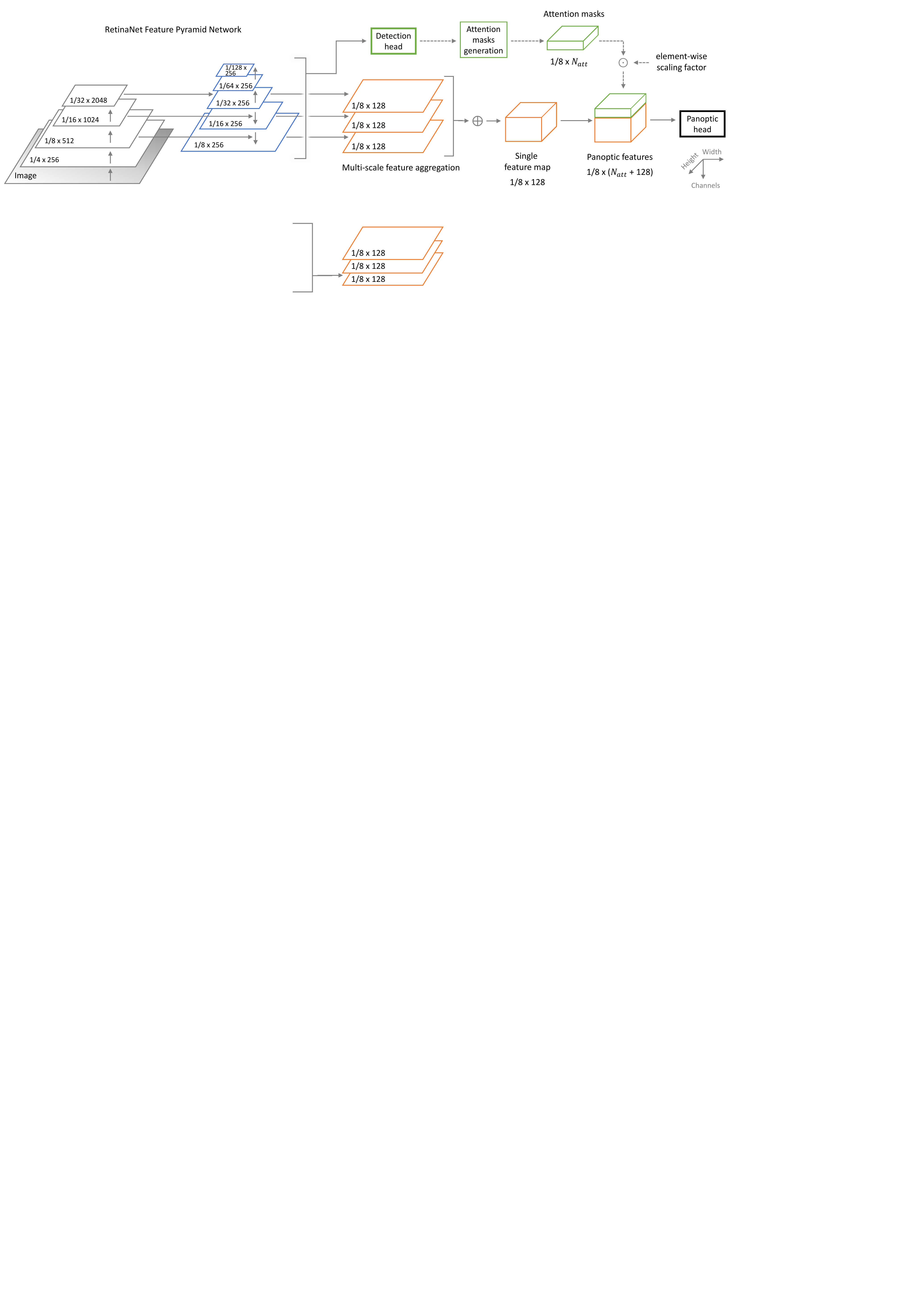}
\caption{Overview of the FPSNet architecture. The dimensions indicate spatial stride on the input image (e.g. $1/8$), and the feature depth (e.g. 128). $\oplus$ denotes element-wise addition. During training, losses are applied only at the two emphasized blocks (detection head and panoptic head). The dotted lines denote that there is no gradient flow in this path during training.}
\label{fig:teaser}
\end{figure*}

Although the existing panoptic segmentation methods achieve state-of-the-art panoptic segmentation quality, there are several drawbacks in terms of speed and computational requirements. Firstly, the merging heuristics are usually performed on CPU, and require looping over all predictions, which is computationally costly.  Secondly, these heuristics need instance masks, and instance segmentation predictions are generally much more computationally expensive and time-consuming than, for example, bounding box object detections. As a result, it is not possible for current methods to make fast panoptic segmentation predictions on high-resolution images, which is desirable for several applications, such as self-driving vehicles and robotics.

To overcome these drawbacks, we present the Fast Panoptic Segmentation Network (FPSNet), an end-to-end architecture that is able to learn how to resolve conflicts between classes and instances. It does not need computationally expensive instance mask predictions or merging operations. Our FPSNet architecture, that is detailed in Section \ref{sec:method}, is compatible with any object detection backbone that is able to generate a single feature map for dense full-image segmentation.

To summarize, we present FPSNet, a fast panoptic segmentation architecture with the following contributions:
\begin{itemize}
    \item FPSNet uses a novel architecture for end-to-end panoptic segmentation that does not require a) instance mask predictions or b) merging heuristics.
    \item With this architecture, we are able to achieve inference speeds much faster than existing methods \cite{Kirillov2019, Xiong2019, Yang2019DeeperLab}, while achieving similar or better Panoptic Quality.
\end{itemize}

In the remainder of this paper, we will first discuss the related work in Section \ref{sec:related_work}. In Section \ref{sec:method}, we define the problem of achieving fast panoptic segmentation, and explain how we solve it with FPSNet. The experiments to evaluate FPSNet are explained in Section \ref{sec:experiments}, and the results are presented in Section \ref{sec:results}. Finally, we provide conclusions in Section \ref{sec:conclusions}.

\section{Related Work}
\label{sec:related_work}

Panoptic segmentation~\cite{Kirillov2018} unifies the typically distinct tasks of semantic segmentation and instance segmentation. Earlier forms of panoptic segmentation have been investigated by many authors~\cite{Arnab2017, yao2012describing, tu2005image}, but only recently it was formulated as a well defined problem by Kirillov \textit{et al.}~\cite{Kirillov2018}. Existing approaches solve this problem either by using separate networks and then fusing the partial results~\cite{Kirillov2018, Li2018ECCV, Arnab2017} or by using a common backbone, and applying a specific head for each subtask followed by late fusion~\cite{Kirillov2019, Li2018AUNet, Xiong2019, Li2018TASCNet, DeGeus2018}.

A baseline solution for panoptic segmentation is given in~\cite{Kirillov2018}, according to which two state-of-the-art networks are trained independently. In that work, the outputs of Mask R-CNN~\cite{He2017} for instance segmentation and PSPNet~\cite{Zhao2017} for semantic segmentation are fused by solving conflicts with heuristics. A clear downside of this method is the fact that it needs two separate networks, which is computationally costly. For this reason, several single network approaches have been presented for the task of panoptic segmentation. 

JSIS-Net~\cite{DeGeus2018}, Panoptic FPN~\cite{Kirillov2019} and the method by Porzi \textit{et al.}~\cite{Porzi2019Seamless} all introduce a common backbone, in order to reduce computation and to benefit from subtasks similarity, and connect it to two separate heads corresponding to the subtasks. TASCNet~\cite{Li2018TASCNet} also consists of a common backbone and two heads, and is augmented with a Things And Stuff Consistency (TASC) loss to enforce per-pixel consistency between the heads. Although this extra loss offers higher consistency in the output distributions, it does not force outputs to be in the panoptic format, and thus heuristics are still needed. AUNet~\cite{Li2018AUNet} applies a single network architecture and extends it with two sources of attention at mask and proposal level, to improve the segmentation of \textit{stuff} classes. To improve the performance for \textit{things} classes, OANet~\cite{Liu2019} proposes a spatial ranking module for solving occlusions between different \textit{things} instances of the same class. Although all single network approaches described above improve computational efficiency by using a single common backbone, they still need to make expensive instance segmentation predictions, and rely on heuristics to generate the final panoptic output.

Another single network approach, UPSNet~\cite{Xiong2019}, applies merging heuristics within the network, and directly outputs the panoptic segmentation predictions, thereby improving the prediction speed. However, UPSNet still makes costly instance segmentation predictions using the two-stage Mask R-CNN~\cite{He2017} method. DeeperLab~\cite{Yang2019} solves the panoptic segmentation problem aiming at efficiency using a single-stage, five-head network, which generates per-pixel semantic and instance predictions. This leads to an efficient neural network, but this method still needs computationally costly merging heuristics to fuse the predictions into a coherent panoptic output. 

The methods discussed so far, all require making instance mask predictions, using merging heuristics, or both. The method by Li \textit{et al.}~\cite{Li2018ECCV}, which is based on the Dynamically Instantiated Network (DIN)~\cite{Arnab2017}, approaches panoptic segmentation in a different way. Here, cues from an external object detector are fused with a semantic segmentation output using a Conditional Random Field (CRF) in order to segment the semantic segmentation output into instances. In earlier work, methods like InstanceCut \cite{Kirillov2017InstanceCut} and the work by Uhrig \textit{et al.}~\cite{uhrig2016pixel} solved the same task with single unified networks, also relying on postprocessing steps to split semantic segmentation predictions into instances. However, they are outperformed by DIN. Compared to the above methods that split predictions instead of merge, FPSNet has the following differences: 1) FPSNet does not make explicit semantic segmentation predictions for \textit{things} classes first. 2) Our method does not need complex post-processing or CRFs to split semantic segmentation outputs into instances. 3) Most importantly, for FPSNet, the entire panoptic segmentation task is learned in an end-to-end fashion and runs at high inference speeds.

In our experiments, we compare with several state-of-the-art approaches including \cite{Kirillov2019, Xiong2019, Porzi2019Seamless, Li2018AUNet, Li2018ECCV}.

\section{Fast Panoptic Segmentation Network}
\label{sec:method}
To achieve fast panoptic segmentation, we aim for a method that does not require:
\begin{enumerate}
    \item making instance segmentation predictions;
    \item a postprocessing step to merge or split predictions.
\end{enumerate}

We achieve this by introducing a novel convolutional neural network module, which we call the \textit{panoptic head}. This head has two inputs: 1) a feature map on which we can perform dense segmentation, and 2) attention masks indicating the presence of \textit{things} instances, and the classes corresponding to those instances, which we obtain from a regular bounding box object detector. From this, the model is trained to 1) perform semantic segmentation for \textit{stuff} classes, 2) morph the attention masks into complete pixel-wise instance masks for \textit{things} instances, and 3) output the predictions for both the \textit{stuff} classes and \textit{things} instances in a single map, on which we can do pixel-wise classification. This module is trained end-to-end in a single network, together with the required feature extractor and bounding box object detector.

We call our network the Fast Panoptic Segmentation Network (FPSNet), and introduce its components in more detail in the next sections. In Section \ref{sec:method:panoptic_module}, we present the novel panoptic module and explain how it is trained. The network backbone is discussed in Section \ref{sec:method:backbone}.

\subsection{Panoptic module}
\label{sec:method:panoptic_module}
In our novel panoptic module for fast panoptic segmentation, we assume that we have bounding box object detections from a regular object detector, as well as a single feature map to apply dense image segmentation. The bounding boxes are used to generate attention masks to indicate the location of \textit{things} in the image, and determine the order of the \textit{things} in the output. The attention masks are first shuffled, then concatenated to the feature map, and finally applied to a fully convolutional network, i.e. the panoptic head. 

At the output of the panoptic head, we predict, for each pixel, either a \textit{stuff} class or a \textit{things} instance \textit{id}, which can directly be related back to a \textit{things} class predicted by the bounding box object detector. Furthermore, at its output, the panoptic head is trained to morph the attention masks into coherent \textit{things} instance masks. In essence, in the output features of the panoptic head, the \textit{stuff} classes and the \textit{things} instance \textit{ids} are treated the same.

\begin{figure}[t]
\centering
\subfigure[Input image with bounding box detections]{\includegraphics[width=0.48\linewidth]{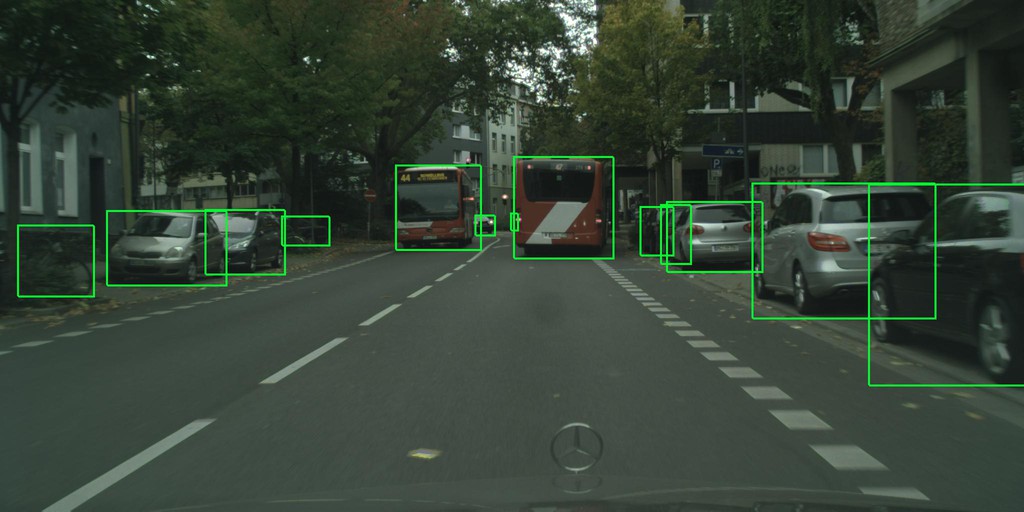}}
\subfigure[Soft attention masks]{\includegraphics[width=0.48\linewidth]{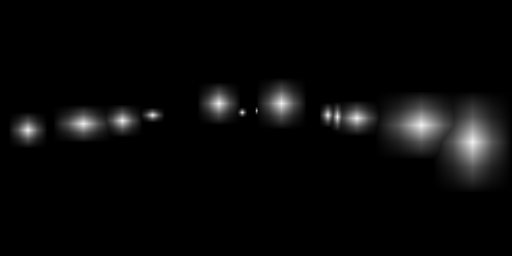}}\\
\subfigure[Color coding shows unique \textit{id} of each attention mask]{\includegraphics[width=0.48\linewidth]{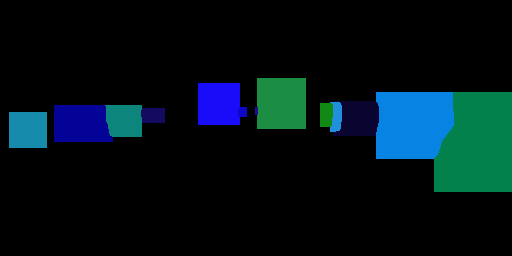}}
\subfigure[Panoptic output]{\includegraphics[width=0.48\linewidth]{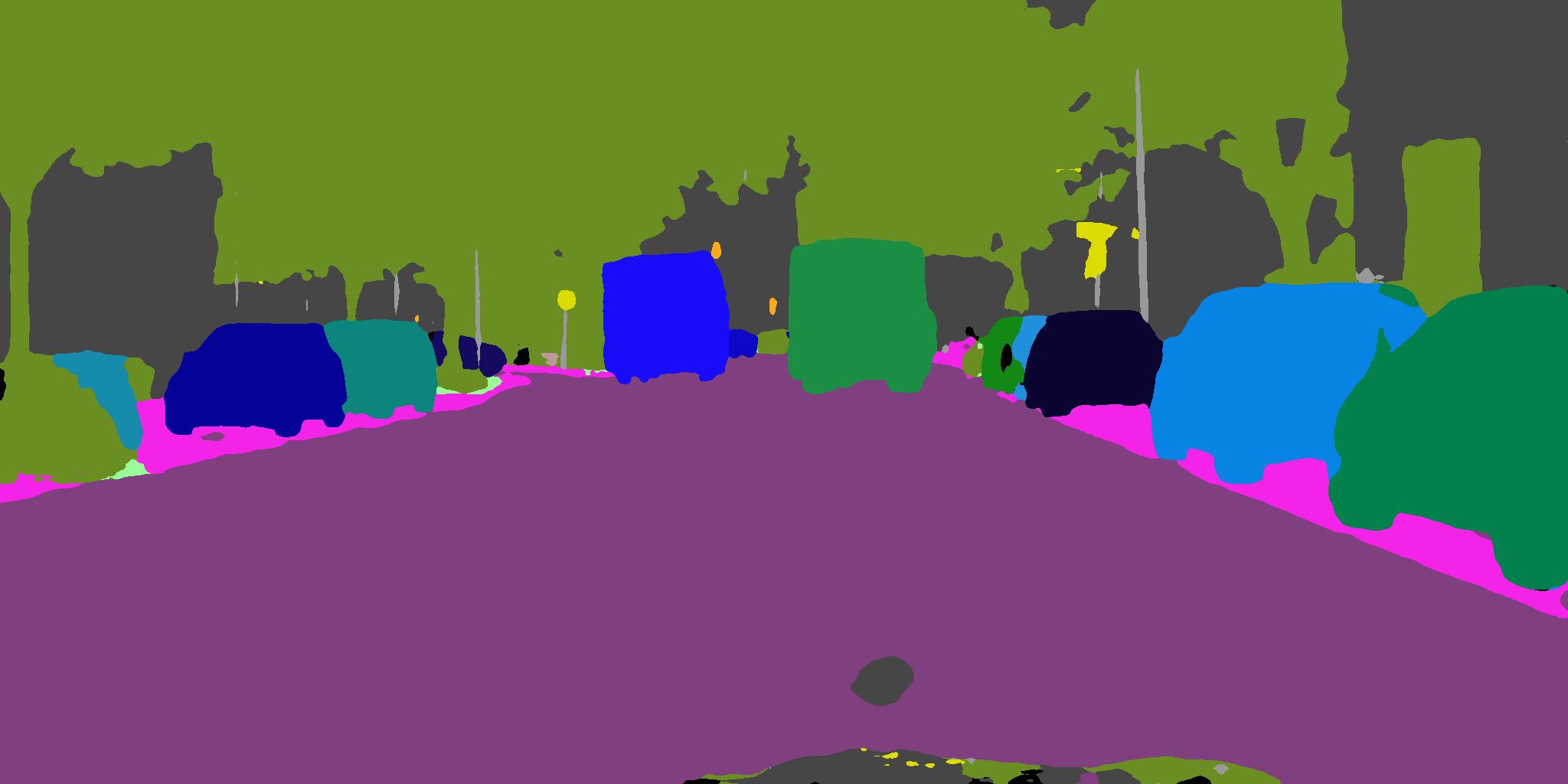}}\\
\caption{FPSNet in steps. Attention masks are generated from the bounding boxes, receive a unique \textit{id} and are used to make panoptic predictions. Image from the Cityscapes dataset.}
\label{fig:results_FPSNet_steps}
\end{figure}

\subsubsection{Attention mask generation}
\label{sec:method:att_mask_generation}
To indicate the location of \textit{things} in the image, we generate attention masks, based on the bounding box object detections. We do this by projecting the bounding boxes on a tensor with the dimensions of the feature map, and filling the bounding box with values of a Gaussian distribution, with mean $\mu = (x_c, y_c)$ and covariance $C = Diag(\frac{1}{4}{w_b}, \frac{1}{4}{h_b})$, where $(x_c, y_c)$, $w_b$, $h_b$ are the center coordinates, width and height of the bounding box, respectively. Outside the bounding boxes, values are 0. This is depicted in Figure \ref{fig:results_FPSNet_steps}. We opt for these so-called \textit{soft} attention masks rather than \textit{hard} masks - with a constant value for all pixels - because we assume that it is more likely that the object is located at the center of the bounding box, and therefore it should receive more attention there. In experiments, we show that soft masks indeed lead to better performance, see Section \ref{sec:experiments:ablations}. In total, we use $N_{att}$ attention masks. If there are more than $N_{att}$ objects detected by the object detector, we pick the $N_{att}$ bounding boxes with the highest class scores. If there are less, we use tensors filled with only zeros for the remaining masks, effectively applying no attention.

After the masks are generated, we shuffle the masks, so that objects of different sizes, class score and class \textit{id} are divided among the channels and filters of the convolutional layers as equally as possible during training. This is done to let each feature dimension related to attention masks to be treated equally as possible. With experiments, we show that mask shuffling boosts the performance (see Section \ref{sec:experiments:ablations}).

\subsubsection{Panoptic head}
\label{sec:method:panoptic_head}
After the attention masks are generated and shuffled, they are stacked in the outer, so-called \textit{channel} dimension of a tensor, so that the tensor has the shape [$N_b$, $H$, $W$, $N_{att}$], where $N_b$ is the batch size, $H$ and $W$ are the height and width of the feature map, respectively, and $N_{att}$ is the number of attention masks. The masks are then scaled so that they are in the range $[0, 1]$, and multiplied with a constant, $C_{att}$, to make sure that the attention masks are in the same order of magnitude as the features, to facilitate learning.

The feature map from the backbone is then concatenated to the attention masks, along the channel dimension, resulting in a tensor with the shape [$N_b$, $H$, $W$, $N_{att}$ + $F_{dim}$], where $F_{dim}$ is the depth of the feature map from the backbone. Subsequently, we apply a 3x3 convolution with ReLU activation~\cite{Nair2010ReLU} and batch normalization~\cite{Ioffe2015BatchNorm}, to merge the concatenated feature map, before feeding it to the head architecture. This head consists of four more 3x3 convolutional layers with ReLU activation and batch normalization.

\begin{figure}
\includegraphics[width=1.0\linewidth]{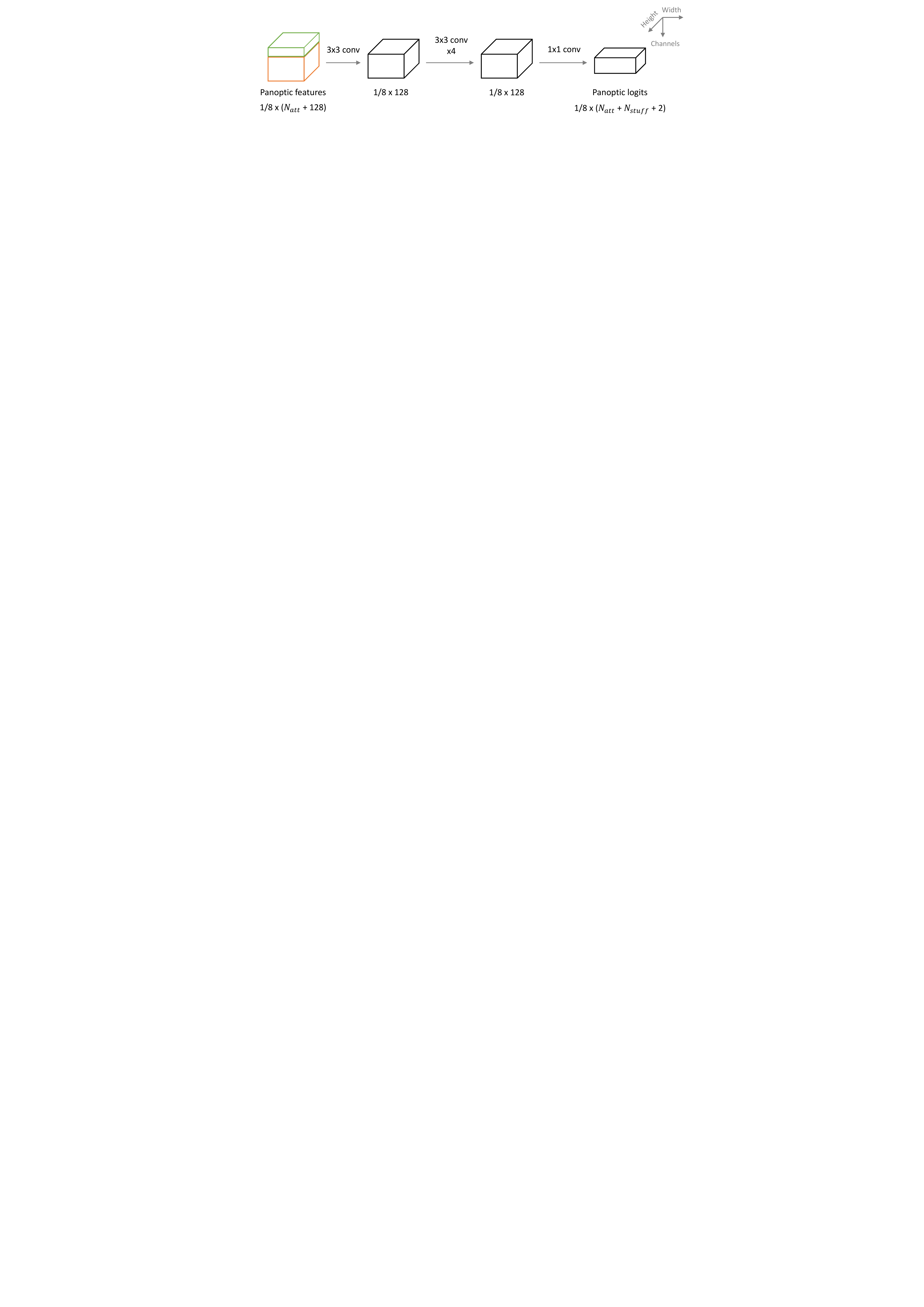}
\caption{The panoptic head architecture.}
\label{fig:panoptic_head}
\end{figure}

To get the final panoptic prediction, we apply a 1x1 convolution to predict $N_{out}$ outputs for each pixel, with $N_{out} = N_{att} + N_{stuff} + 2$. Here, $N_{stuff}$ is the number of \textit{stuff} classes. The pixels that are predicted for the first $N_{att}$ outputs can be related back to the input attention masks. The pixels for the $n$th output belong to the $n$th attention mask (after shuffling) and its corresponding class. For each pixel, we get the final panoptic prediction by picking the instance or class with the highest score, i.e. applying \textit{argmax}, after bilinearly upsampling the logits to the dimensions of the input image. The panoptic head architecture is depicted in Figure \ref{fig:panoptic_head}.

\subsubsection{Training}
\label{sec:method:training}
Because panoptic segmentation only allows a single prediction for each pixel, we treat the problem as a semantic segmentation problem during training. We construct a ground-truth consisting of a single prediction for each pixel, and apply a softmax cross-entropy loss. The desired output for each pixel is either a \textit{stuff} class or a \textit{things} instance \textit{id}, keeping in consideration that it is the responsibility of the object detector to provide the class label for each \textit{things} instance.

The main challenge is to make sure that the order of the feature dimensions in the output tensor of the panoptic head related to \textit{things} instance \textit{ids}, is the same as the order in which the attention masks are stacked in the input tensor of the panoptic head. Otherwise, it is not possible to relate a feature dimension in the output back to a \textit{things} class of the object detector. We achieve this by matching the predicted attention masks to the ground-truth \textit{things} instances, and simply re-ordering the matched ground-truth \textit{things} instances such that their order corresponds to the order of the attention masks in the input tensor of the panoptic head. When training the panoptic head, we assume that we have accurate attention masks. This means that, during training, we only assign one single predicted attention mask to a ground-truth instance, and vice versa. Thus, after the attention masks are gathered, we discard the ones that do not have an Intersection over Union (IoU) greater than 0.5 with a ground-truth instance, and we assign each \textit{things} instance only to the attention mask for which the IoU is highest.

The supervision for the \textit{stuff} classes is the same as for a semantic segmentation problem. In our case, the \textit{stuff} ground-truth is concatenated to the \textit{things} ground-truth instance masks. The pixels of the unmatched \textit{things} instances and the \textit{unlabeled} pixels are the second to last and the last entries of the ground-truth tensor, respectively.

By constructing the loss and ground-truth in this fashion, the network learns to output the \textit{things} instances in the same order as the input attention masks. 
An example is shown in Figure \ref{fig:results_FPSNet_steps}. Note that we effectively apply class-agnostic instance segmentation using these attention masks, and that the relevant \textit{things} classes are retrieved from the object detector using the order-preserving nature of the panoptic head.

\subsection{Backbone}
\label{sec:method:backbone}
For the FPSNet framework, we need a backbone that performs object detection and is able to generate a single feature map. The single feature map is necessary to make dense panoptic segmentation predictions, and the object detection network is required to output the bounding boxes that are used to generate attention masks for \textit{things} instances.

\subsubsection{Object detection network}
Since FPSNet can work with various types of object detectors, and we aim to achieve fast, but also accurate, panoptic segmentation, we pick RetinaNet \cite{Lin2017RetinaNet} as our object detection network. RetinaNet is a single-stage object detector that achieves state-of-the-art performance at high inference speeds. In our implementation, we use the version of RetinaNet with a ResNet-50-based Feature Pyramid Network (FPN) as backbone \cite{He2015, Lin2017, Lin2017RetinaNet}. This can be seen in Figure \ref{fig:teaser}.

\begin{table*}[t]

\centering
\caption{Speed and accuracy results for FPSNet and other methods on the Cityscapes validation set. Values are as reported in the respective papers, unless indicated otherwise. (L)W-MNV2 is (Light) Wider MobileNetV2 \cite{Yang2019DeeperLab}.}
\begin{tabular}{ l || c | c || c || c | c | c }
\multicolumn{1}{c||}{} & \multicolumn{1}{c |}{} & \multicolumn{1}{c ||}{} & \multicolumn{1}{c ||}{\textit{Accuracy}} & \multicolumn{3}{c}{\textit{Prediction time}} \\
 Method & Backbone & Resolution & PQ (\%) $\uparrow$ & Inference (ms) $\downarrow$ & Merging (ms) $\downarrow$ & Total (ms) $\downarrow$ \\ \hline
DeeperLab \cite{Yang2019DeeperLab} & LW-MNV2 & 512 x 1024 & 39.2 & 41 & 45 & 86 \\ 
FPSNet (ours) & ResNet-50-FPN & 512 x 1024 & \textbf{46.7} & 45 & \textit{n/a} & \textbf{45} \\ \hline
DeeperLab \cite{Yang2019DeeperLab} & LW-MNV2 & 1024 x 2048 & 48.1 & 97 & 154 & 251  \\
DeeperLab \cite{Yang2019DeeperLab} & W-MNV2 & 1024 x 2048 & 52.3 & 149 & 154 & 303  \\
DeeperLab \cite{Yang2019DeeperLab} & Xception-71 & 1024 x 2048 & 56.5 & 308 & 154 & 462 \\
UPSNet \cite{Xiong2019} & ResNet-50-FPN & 1024 x 2048 & \textbf{59.3} & 202 & \textit{n/a} & 202 \\
Panoptic FPN (our re-impl.) & ResNet-50-FPN & 1024 x 2048 & 57.0 & 226 & 152 & 378 \\
FPSNet (ours) & ResNet-50-FPN & 1024 x 2048 & 55.1 & 114 & \textit{n/a}& \textbf{114} \\

\end{tabular} 
\label{tab:results_main_cityscapes}
\end{table*}

\begin{table}[t]

\centering

\caption{Main results for FPSNet on the Cityscapes validation set.}
\begin{tabular}{ l || c ||@{\hspace{4px}} c @{\hspace{4px}}|@{\hspace{4px}} c @{\hspace{4px}}|@{\hspace{4px}} c @{\hspace{4px}}| @{\hspace{4px}}c @{\hspace{4px}}}
 Method & Backbone & PQ & PQ\textsubscript{Th} & PQ\textsubscript{St} & Time (ms)\\ \hline
De Geus \textit{et al.}~\cite{DeGeus2019} & RN-50  & 45.9 & 39.2 & 50.8 & - \\ 
Li \textit{et al.}~\cite{Li2018ECCV} & RN-101 & 47.3 & 39.6 & 52.9 & -  \\
TASCNet \cite{Li2018TASCNet} & RN-50-FPN & 55.9 & 50.5 & 59.8 & -  \\
DeeperLab \cite{Yang2019DeeperLab} & XC-71 & 56.3 & 52.7 & 59.0  & 462 \\
AUNet \cite{Li2018AUNet} & RN-50-FPN & 56.4 & 52.7 & 59.0  & - \\
Panoptic FPN  \cite{Kirillov2019} & RN-50-FPN & 57.7 & 51.6 & 62.2 & - \\
UPSNet \cite{Xiong2019} & RN-50-FPN & 59.3 & 54.6 & 62.7 & 202 \\
Porzi \textit{et al.}~\cite{Porzi2019Seamless} & RN-50-FPN & \textbf{60.2} & \textbf{55.6} & \textbf{63.6} & - \\ \hline
FPSNet (ours) & RN-50-FPN & 55.1 & 48.3 & 60.1 & \textbf{114} \\
\end{tabular} 
\label{tab:results_pq_cityscapes}
\end{table}

\subsubsection{Single feature map}
The output of the ResNet-50-based Feature Pyramid Network is a set of feature maps from different levels of the feature extractor. However, to make dense panoptic segmentation predictions, we need a single feature map. In \cite{Kirillov2019}, the authors encountered a similar problem of performing the task of semantic segmentation on a multi-scale feature map. They solved this by upsampling and merging the different layers of the feature map, to finally generate a single feature map. For our implementation, we maintain a similar approach. As seen in Figure \ref{fig:teaser}, the output of the FPN is a set of feature maps \{P3, P4, P5, P6, P7\}, with strides \{8, 16, 32, 64, 128\}, respectively. We use feature maps P3, P4, and P5 to generate our single feature map, with a stride of 8. We apply two upsampling steps to P5 and one upsampling step to P4, creating S5 and S4, respectively. Each upsampling step consists of a 3x3 convolutional layer with ReLU, followed by $2\times$ bilinear upsampling. We get S3 by applying a 3x3 convolutional layer with ReLU to P3. Finally, we generate the final feature map S with ${S} = {S3} + {S4} + {S5}$. Note that this is very similar to the process maintained in Panoptic FPN, except for the fact that we do not use a feature map with a stride of 4, to save computation time and resources. RetinaNet \cite{Lin2017RetinaNet} applies a similar strategy to achieve more efficient object detection than FPN \cite{Lin2017}.

\subsubsection{Training}
The object detection head of the network is trained in the usual fashion, as explained in \cite{Lin2017RetinaNet}.

\section{Experiments}
\label{sec:experiments}
We conduct the following experiments to demonstrate FPSNet and evaluate its performance:
\begin{itemize}
    \item \textit{Speed and accuracy}: Since FPSNet is designed for both speed and accuracy, we evaluate both at different resolutions, and compare with existing methods. For these experiments, we use the Cityscapes dataset \cite{Cordts2016Cityscapes}.
    \item \textit{Ablation study}: We conduct ablation experiments to show the effect of various design choices, i.e. attention mask shuffling, the use of hard attention masks, and tuning $N_{att}$ and $C_{att}$. Again, we evaluate on the Cityscapes dataset.
    \item \textit{Performance on Pascal VOC}: To demonstrate the general applicability of FPSNet, we evaluate on the Pascal VOC dataset \cite{Everingham2010Pascal}.
\end{itemize}

\subsection{Metrics}
We evaluate the performance of our panoptic segmentation method using the Panoptic Quality (PQ) metric \cite{Kirillov2018}. This metric includes both the recognition and segmentation capabilities of the network. We also assess the performance of our network for \textit{things} and \textit{stuff} classes separately, through PQ\textsubscript{Th} and PQ\textsubscript{St}, respectively.

To assess the prediction speed of the network, we also measure its inference time of the network. Since FPSNet does not need additional postprocessing to generate panoptic segmentation predictions, the total time required for a prediction is directly given by the network inference time. We report single image inference time on an Nvidia Titan RTX GPU, averaged over all images in the validation set.

\subsection{Datasets}
We evaluate FPSNet on Cityscapes \cite{Cordts2016Cityscapes}. Cityscapes is a specialized dataset consisting of 5k street scene images. It has annotations for 8 \textit{things} and 11 \textit{stuff} classes. To prevent overfitting, we apply a data augmentation strategy similar to the one described in \cite{Kirillov2019}. We randomly scale the image with a random factor between 0.5 and 1.5, and use a random crop of 512x1024 pixels as input to the network.

To test the applicability of FPSNet on other datasets, we also train and test on Pascal VOC \cite{Everingham2010Pascal}. Pascal VOC is a more general computer vision dataset. As in other related work \cite{Yang2019DeeperLab, Li2018ECCV}, we generate a training set by merging the Pascal VOC 2012 training set and the additional annotations from the SBD dataset \cite{Bharath2011SBD}. This results in 10582 training images. For validation, we use the Pascal VOC 2012 validation set. This dataset has annotations for 20 \textit{things} classes and no \textit{stuff} classes. We randomly resize the images to square images between 512x512 and 800x800 pixels, and train on random crops of 512x512 pixels.

\subsection{Implementation details}
\label{sec:experiments:implementation}
Since FPSNet applies both object detection and panoptic segmentation, the loss function is given by

\begin{equation}
\label{eq:loss}
{L} = \lambda_{det}L_{det} + \lambda_{pan}L_{pan},
\end{equation}

where $L_{det}$ are the RetinaNet detection losses defined in \cite{Lin2017RetinaNet}, $L_{pan}$ is our softmax cross-entropy loss for panoptic segmentation, and $\lambda_{det}$ and $\lambda_{pan}$ are the respective loss weights. In our implementation, $\lambda_{det} = 0.5$ and $\lambda_{pan} = 1.0$, as we found that this led to the best results. We train our network by optimizing the loss using stochastic gradient descent with a momentum of 0.9. The weight decay is 0.001. We train all networks on a single GPU, with a batch size of 4 images. We use polynomial learning rate schedule (as in \cite{Chen2018}) with an initial learning rate of 0.01 and a power of 0.9. For the main \textit{speed and accuracy} experiments, we train the network for 200k steps. For the ablation experiments, we train all networks for 100k steps. By default, we use $N_{att} = 50$ and $C_{att} = 50$; ablations are provided in Section \ref{sec:experiments:ablations}. Before training, the backbone is initialized with weights from a model pre-trained on ImageNet~\cite{Deng2009ImageNet}.

\section{Results}
\label{sec:results}
\subsection{Speed and accuracy}
\label{sec:results:main}
\begin{figure}[t]
\centering
\includegraphics[width=0.480\linewidth]{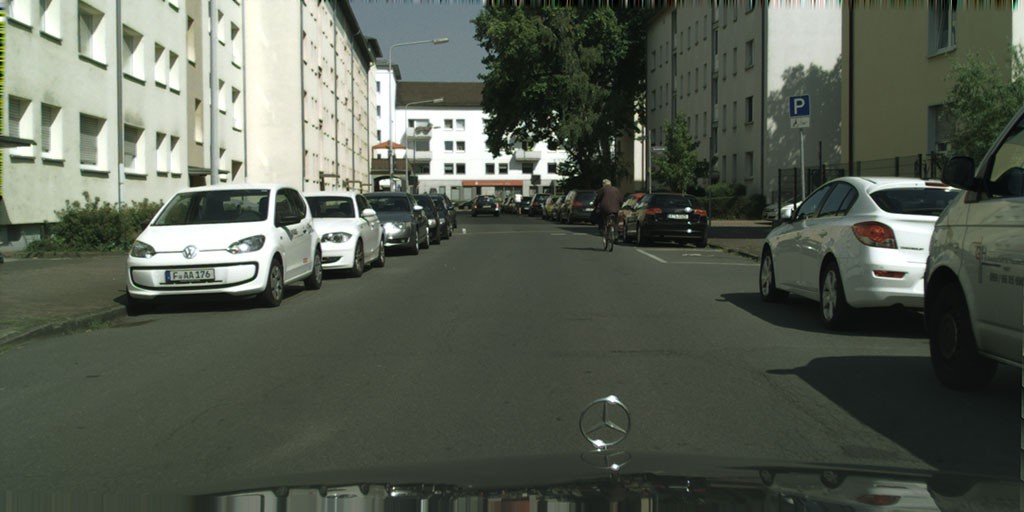}
\includegraphics[width=0.480\linewidth]{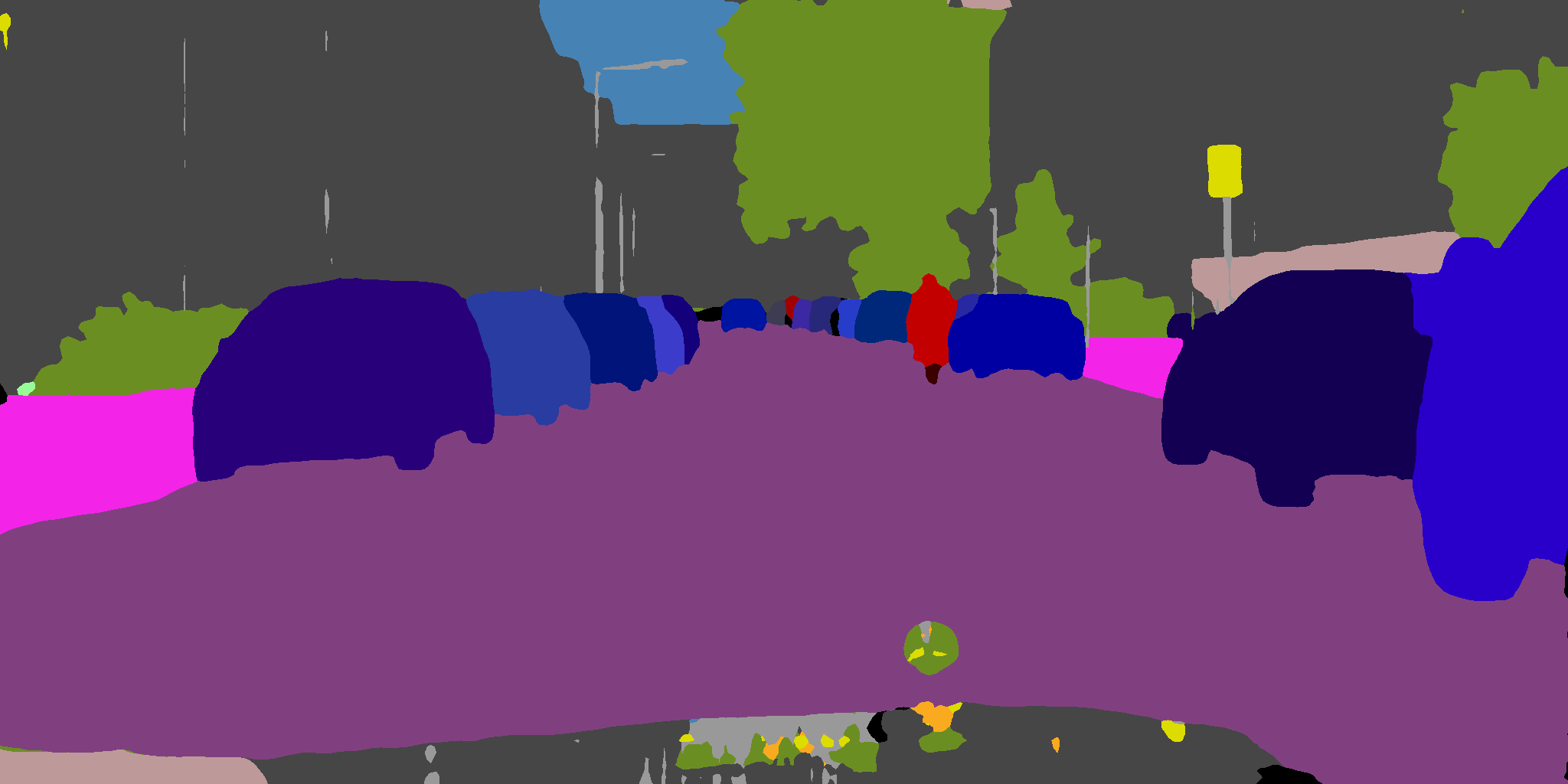}\\
\includegraphics[width=0.480\linewidth]{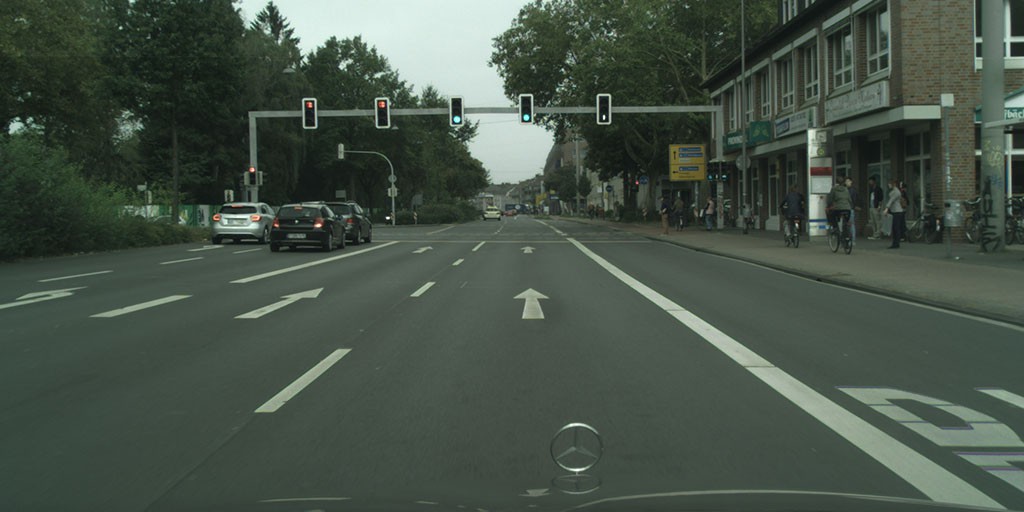}
\includegraphics[width=0.480\linewidth]{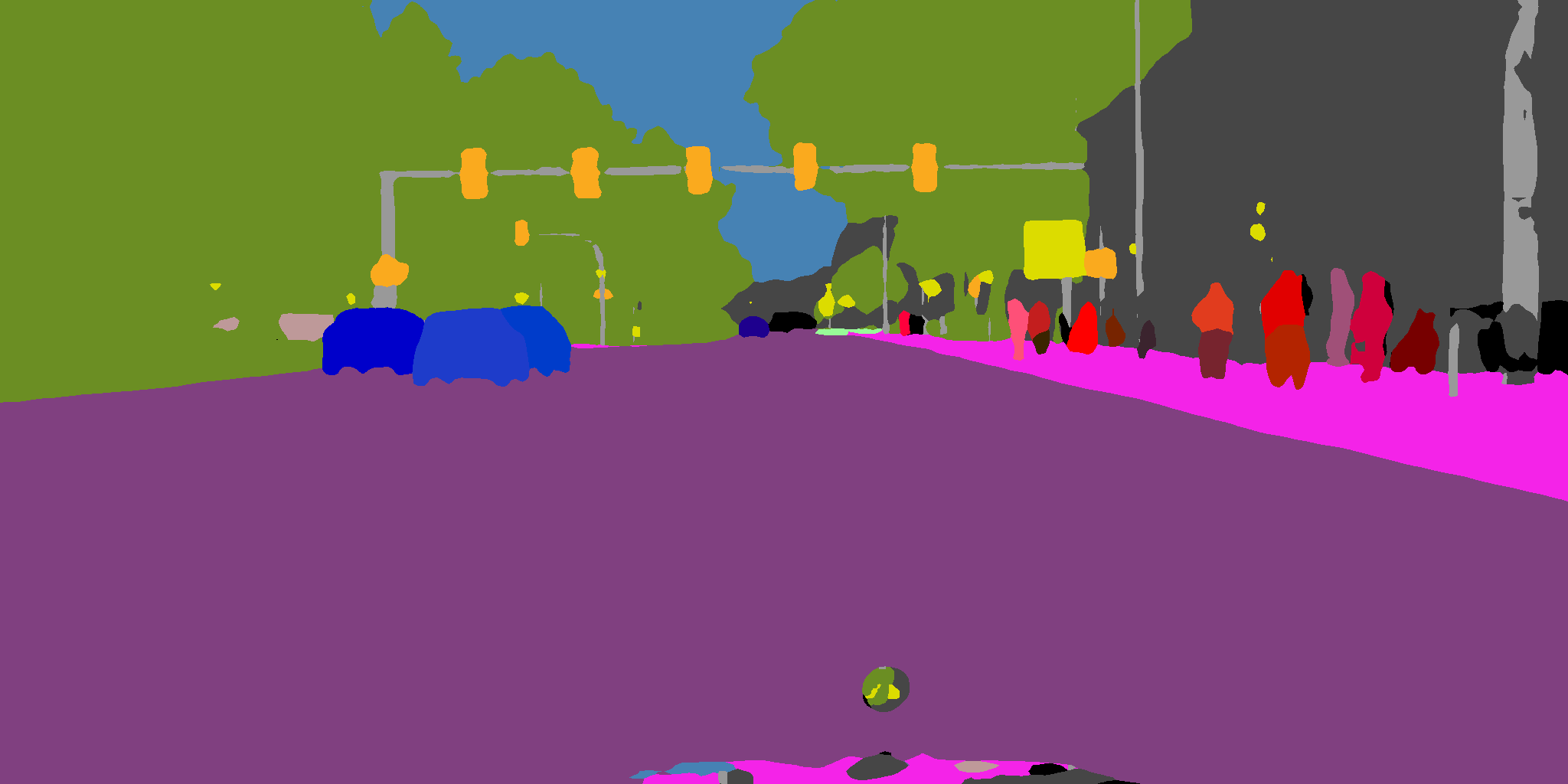}\\
\includegraphics[width=0.480\linewidth]{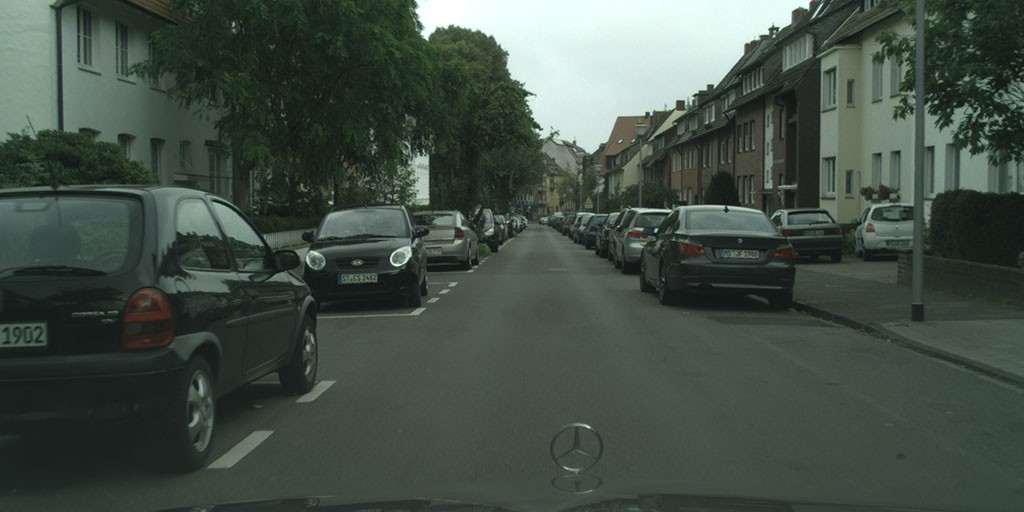}
\includegraphics[width=0.480\linewidth]{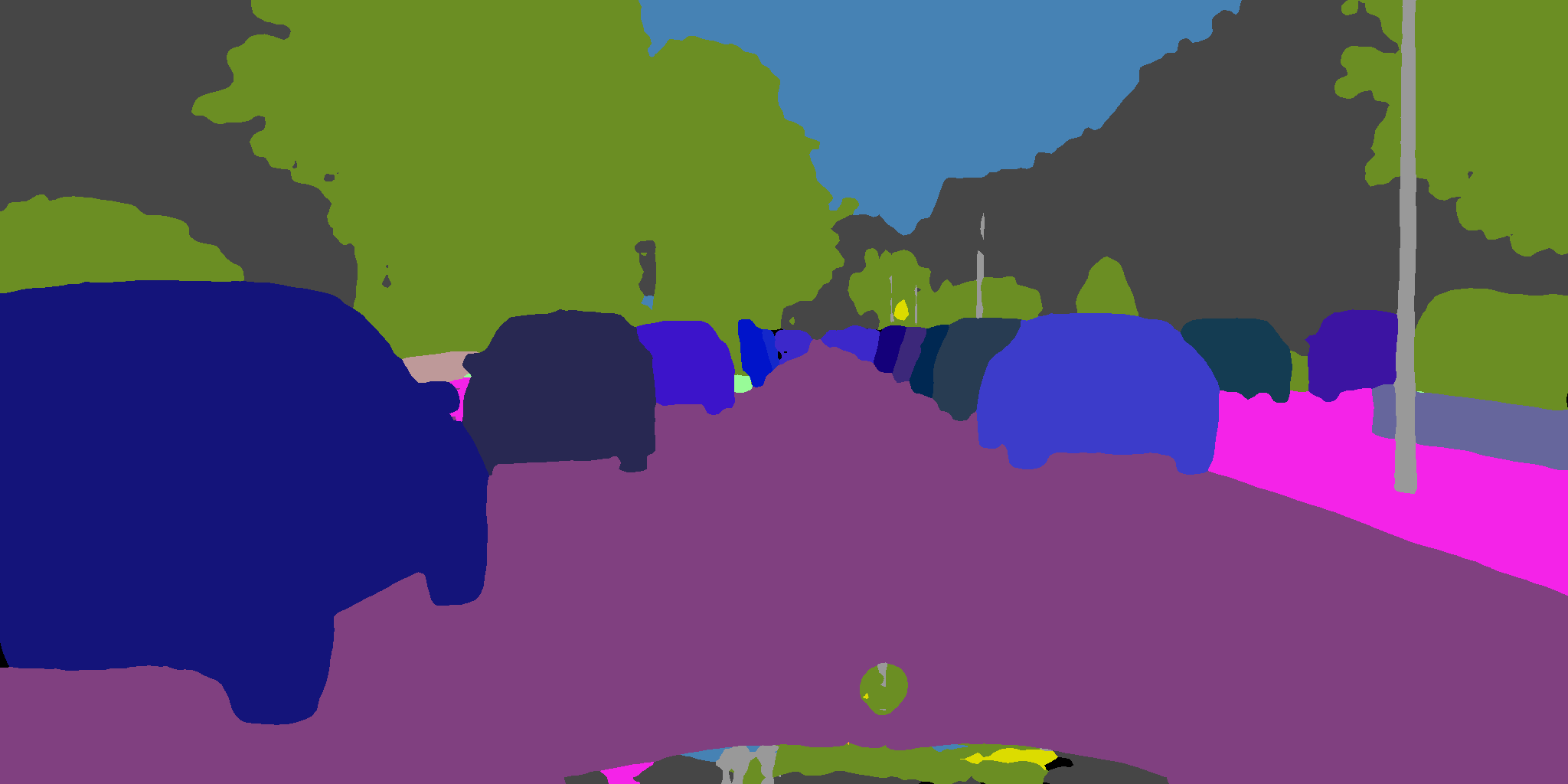}\\
\includegraphics[width=0.480\linewidth]{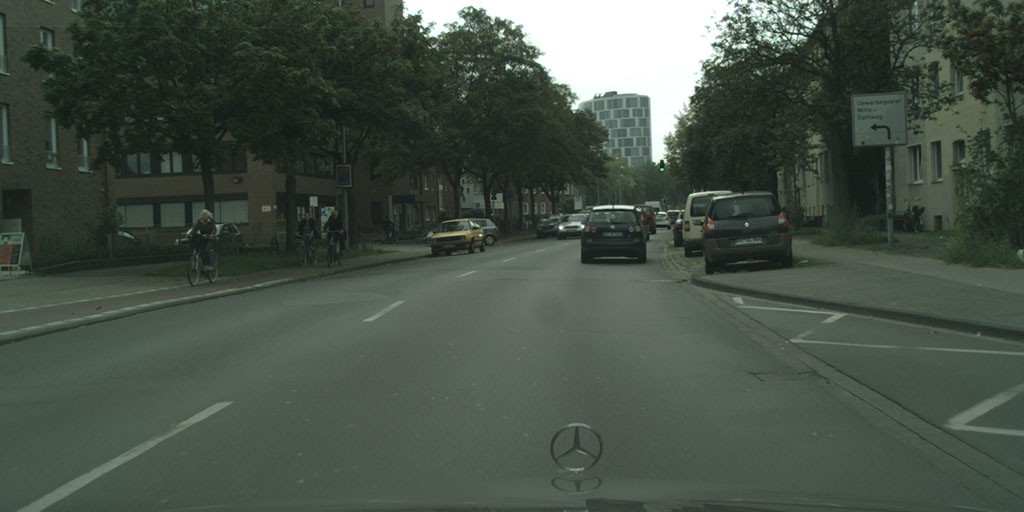}
\includegraphics[width=0.480\linewidth]{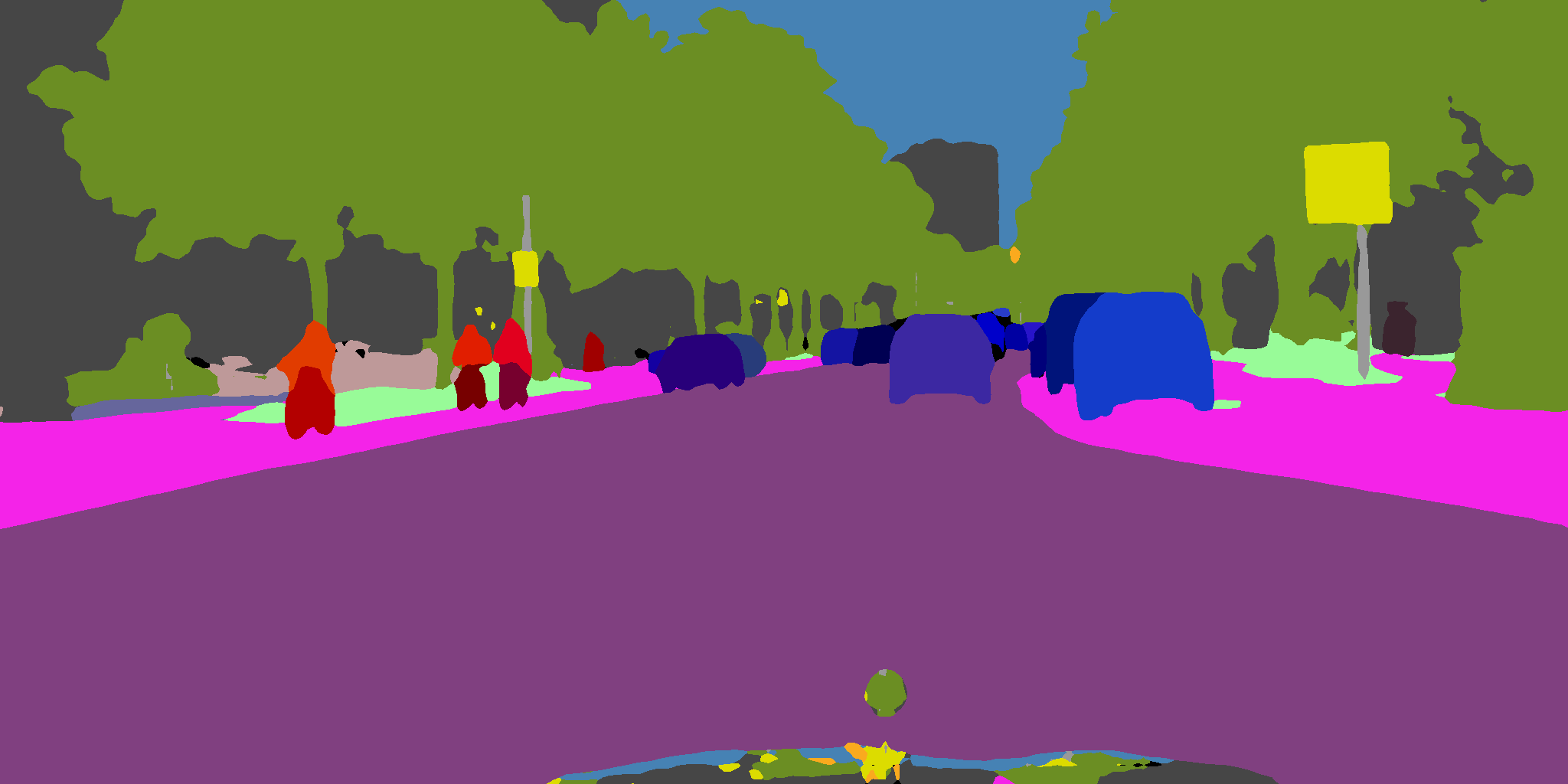}\\
\includegraphics[width=0.480\linewidth]{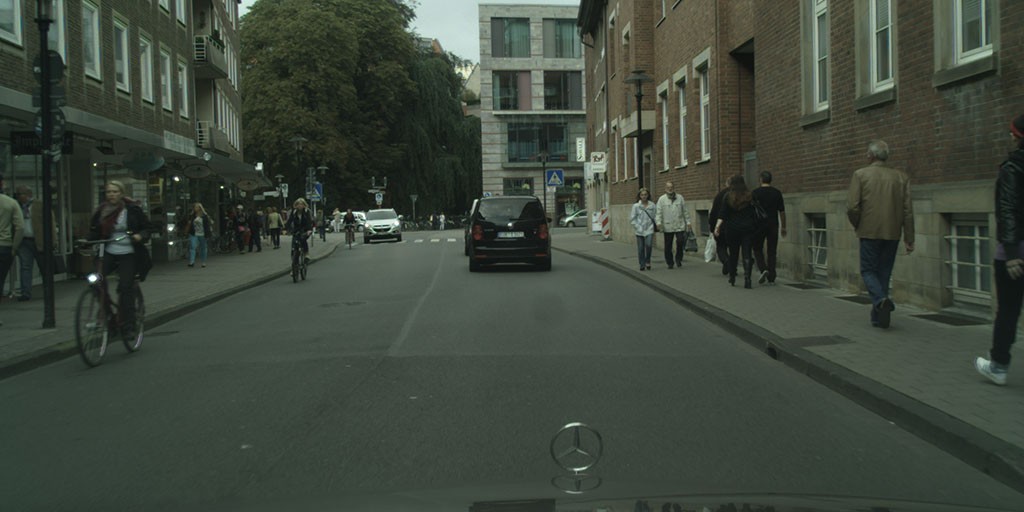}
\includegraphics[width=0.480\linewidth]{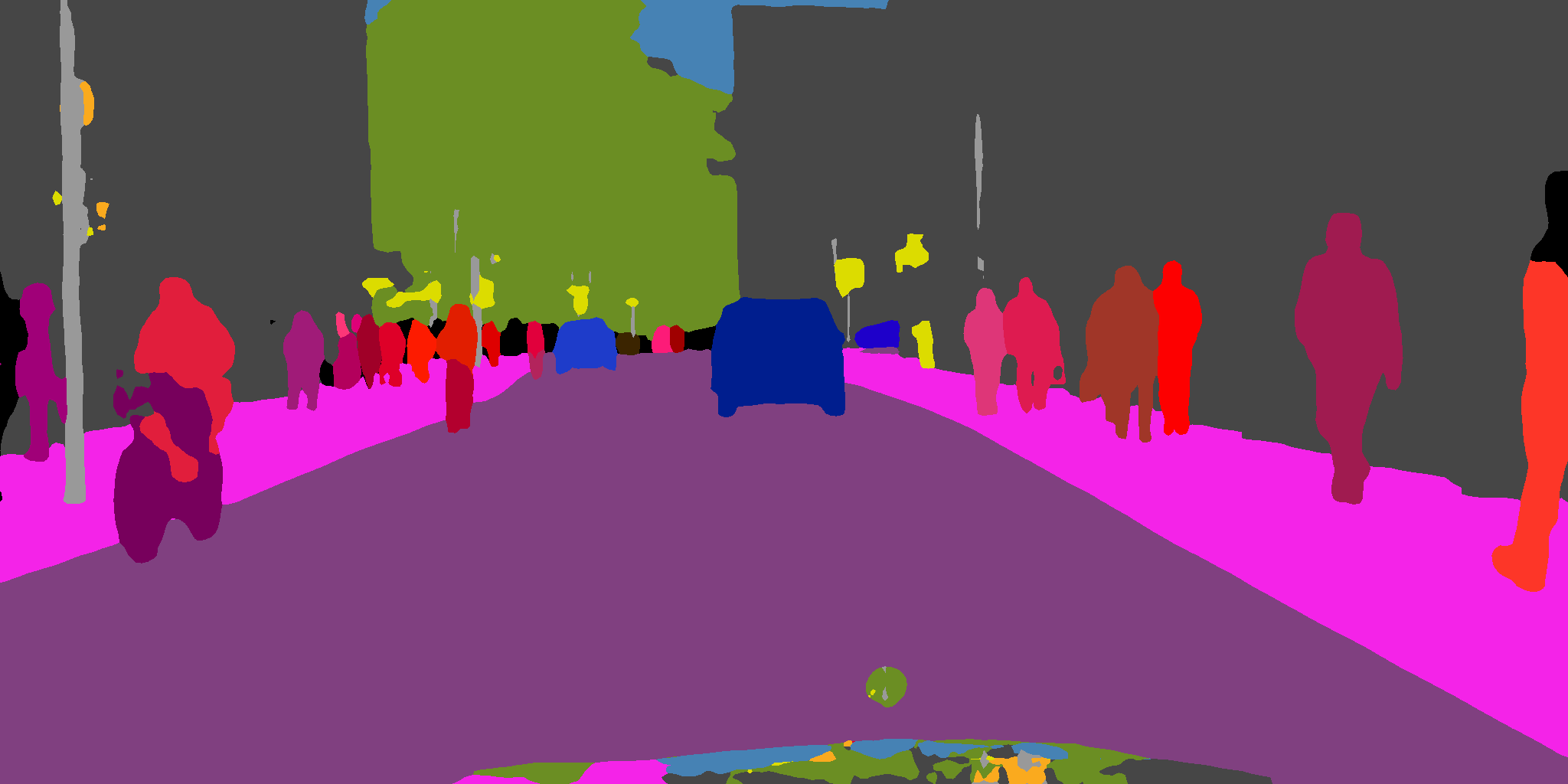}\\
\includegraphics[width=0.480\linewidth]{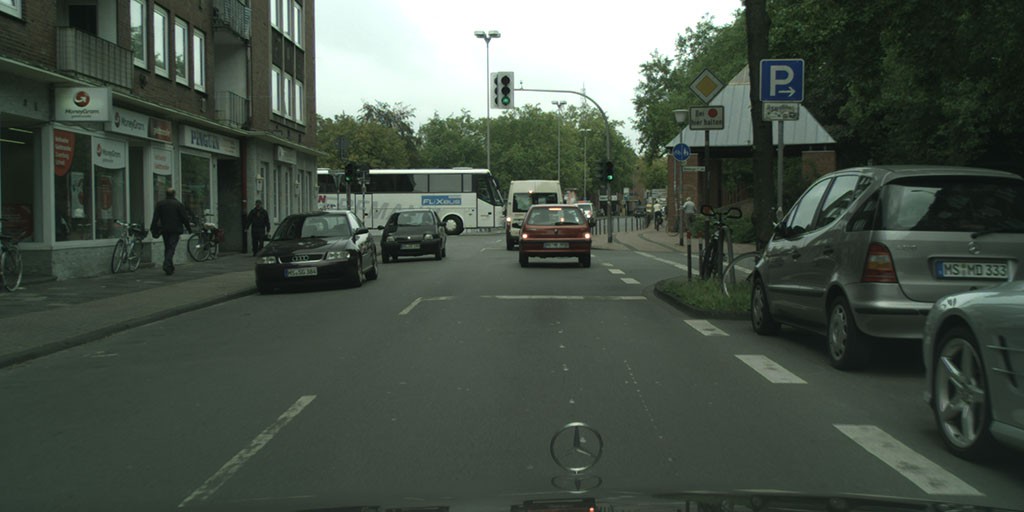}
\includegraphics[width=0.480\linewidth]{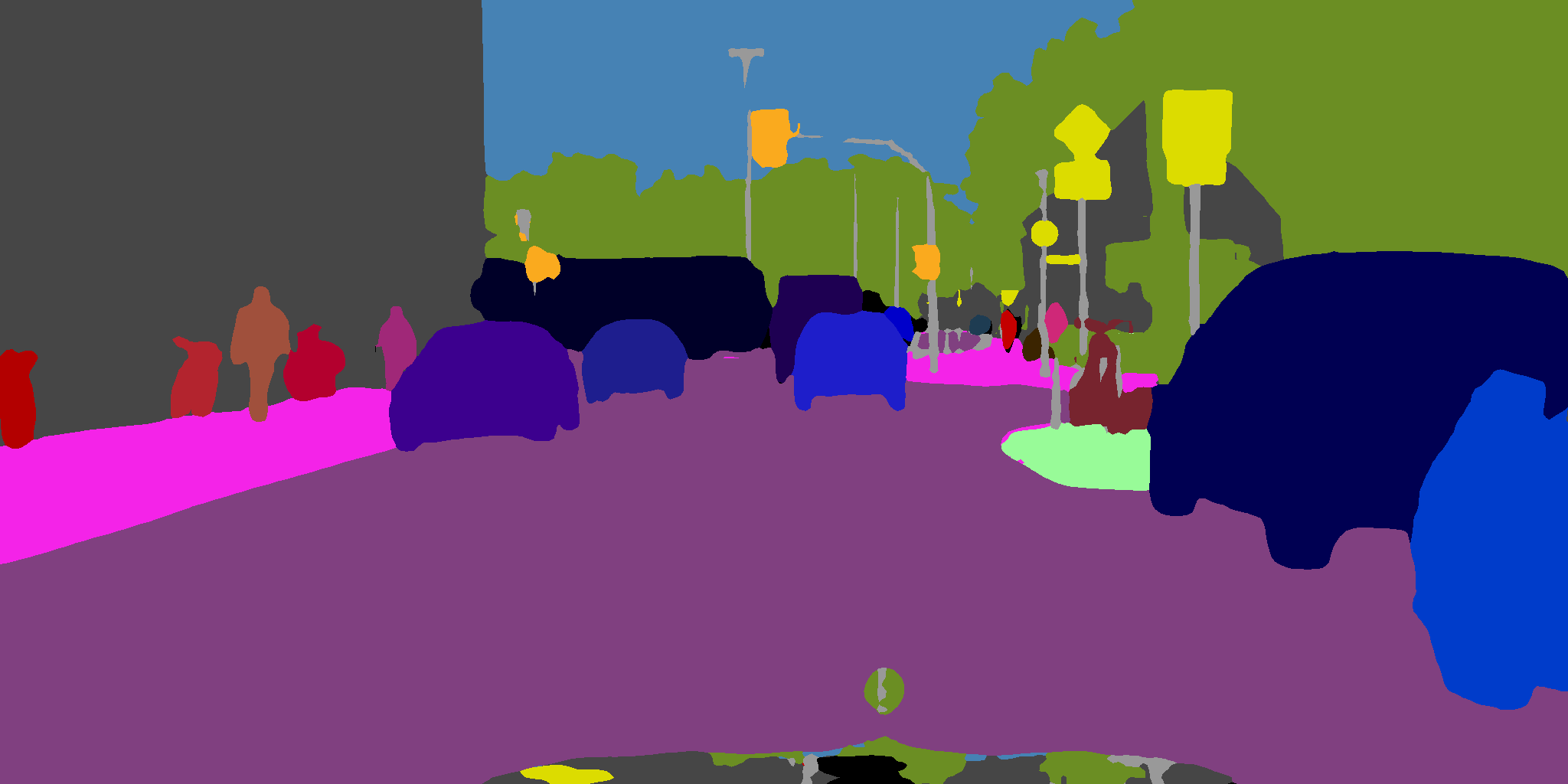}
\caption{Example predictions by FPSNet on the Cityscapes validation set. Each color indicates a different \textit{things} instance or \textit{stuff} class.}
\label{fig:results_examples}

\end{figure}

In Table \ref{tab:results_main_cityscapes}, we present PQ scores and prediction times for FPSNet and existing methods that report prediction times. Unless indicated otherwise, all scores and prediction times are as reported in the respective papers. From Table \ref{tab:results_main_cityscapes}, it follows that FPSNet is considerably faster than existing panoptic segmentation methods, while still achieving competitive scores on Panoptic Quality. Comparing with DeeperLab \cite{Yang2019DeeperLab}, a panoptic segmentation method designed for speed and efficiency, it becomes clear that FPSNet achieves higher PQ scores at lower inference times. At a PQ score of 52.3, DeeperLab is almost three times slower than FPSNet at a PQ score of 55.1. 
UPSNet \cite{Xiong2019} does score significantly higher than FPSNet, but it is also two times slower than our slowest implementation. In Figure \ref{fig:time_pq_plot}, the different prediction times and Panoptic Quality scores are visualized. Qualitative results on the Cityscapes validation set are shown in Figures \ref{fig:main_pred} and \ref{fig:results_examples}.

Additionally, we also compare FPSNet to our own re-implementation of Panoptic FPN \cite{Kirillov2019}. Based on inference time alone, FPSNet is almost $2\times$ as fast. When we take the merging operations into account as well, our method becomes over $3\times$ faster Panoptic FPN.

\begin{table}[t]
\centering
\caption{Several ablations on Cityscapes validation set.}
\begin{tabular}{ c | c | c || c | c | c }
Attention & Hard & GT &   &   &  \\ 
Mask & Attention & Bounding & PQ & PQ\textsubscript{Th} & PQ\textsubscript{St} \\
Shuffling & Masks & Boxes &   &   &  \\
\hline
- & - & - & 53.1 & 44.7 & 59.4 \\ 
\checkmark & - & - & \textbf{54.1} & \textbf{46.7} & \textbf{59.5} \\ 
\checkmark & \checkmark  & - & 52.7  & 43.7 & 59.2 \\ \hline
- & - & \checkmark & 50.5 & 39.7 & 58.4 \\
\checkmark & - & \checkmark & \textbf{57.5} & \textbf{54.7} & \textbf{59.5} \\
\checkmark & \checkmark & \checkmark & 56.1 & 51.5 & 59.4 \\
\end{tabular} 

\label{tab:results_ablation_masks}
\end{table}

\begin{table}[t]
\centering
\caption{Ablations for N\textsubscript{att} and C\textsubscript{att} on Cityscapes validation set.}
\begin{tabular}{ c | c || c | c | c}
N\textsubscript{att} & C\textsubscript{att} & PQ & PQ\textsubscript{Th} & PQ\textsubscript{St}\\ \hline
50 & 50 & \textbf{54.1} & \textbf{46.7} & \textbf{59.5}\\ \hline
25 & 50 & 53.7 & 45.8 & 59.5 \\ 
100 & 50 & 52.6 & 44.5 & 58.5 \\ \hline
50 & 1 & 52.6 & 43.9 & 58.9 \\
50 & 25 & 53.7 & 46.2 & 59.2 \\
50 & 100 & 53.2 & 45.7 & 58.6 \\
\end{tabular} 
\label{tab:results_ablation_natt_catt}
\end{table}

In Table \ref{tab:results_pq_cityscapes}, we compare our performance with a wide range of state-of-the-art panoptic segmentation methods. We compare with methods also using ImageNet initialization and similar backbones. From this, it becomes clear that, even though the focus of FPSNet is on fast panoptic segmentation, we still achieve competitive results on Panoptic Quality. It should be noted that, even though no prediction times are provided for some methods in Table \ref{tab:results_pq_cityscapes}, all of these methods -- except Li \textit{et al.} \cite{Li2018ECCV} -- require instance mask predictions and costly merging operations.
The method presented by Li \textit{et al.} \cite{Li2018ECCV} consists of multiple separate networks, which makes this method inefficient.

\subsection{Ablation study}
We conduct several ablation experiments on the Cityscapes validation set. We evaluate the method using both the original attention masks, gathered from the detection branch output, and attention masks generated using ground-truth bounding boxes. We use these ground-truth bounding boxes for a fair analysis of the performance of the panoptic head.

\label{sec:experiments:ablations}
\subsubsection{Attention mask shuffling} 
To show that attention mask shuffling boosts the performance, we conduct an experiment without shuffling. From Table \ref{tab:results_ablation_masks}, it follows that the performance for \textit{things} classes increases when we introduce attention mask shuffling. The gap is 2 points when the original attention masks are used, but increases to 15 points when using ground-truth bounding boxes as attention masks. In the latter case, there are more attention masks, and they are ordered differently. This shows that not all filters of the convolutional layers in the panoptic head receive adequate supervision, and that the network learns a specific order of \textit{things} instances instead.

\subsubsection{Hard attention masks} 
 We replace our soft attention masks with hard attention masks, where all pixels within the bounding box get the value $C_{att}$. As expected, the results in Table \ref{tab:results_ablation_masks} show that using hard attention masks reduces the performance for \textit{things} classes, given that PQ\textsubscript{Th} is reduced by 3.0 points.

\subsubsection{N\textsubscript{att} and C\textsubscript{att}} 
We train FPSNet with different values for $N_{att}$ and $C_{att}$ and report the results in Table \ref{tab:results_ablation_natt_catt}. We find that changing the number of attention masks, i.e. $N_{att}$, has a slight effect on the performance. Using 25 attention masks instead of 50 slightly decreases the scores on \textit{things} classes, which is caused by a lower performance on images with more than 25 \textit{things} instances. With $N_{att} = 100$, the performance drops for both \textit{stuff} and \textit{things} classes. This is to be expected, since there are more outputs for the final convolutional layer, meaning that it effectively has to learn more. This makes the task more complex and leads to lower performance. From the results, it also follows that $C_{att} = 50$ seems to be the optimal value. In this case, there is a good balance between the magnitude of the attention mask tensors and the features from the feature map. It is likely that there are better ways to create this balance, e.g. with various normalization techniques, but we leave this for future work to address.

\begin{table}[t]

\centering

\caption{Main results for FPSNet on the Pascal VOC validation set. 
}
\begin{tabular}{ l ||@{\hspace{4px}} c @{\hspace{4px}}|@{\hspace{4px}} c @{\hspace{4px}}||@{\hspace{4px}} c @{\hspace{4px}}|@{\hspace{4px}} c @{\hspace{4px}}}
 Method & Backbone & Resolution & PQ  (\%) & Total time (ms) \\ \hline
DeeperLab \cite{Yang2019DeeperLab} & LW-MNV2 & 512 x 512 & 54.1 & 47  \\
DeeperLab \cite{Yang2019DeeperLab} & W-MNV2 & 512 x 512 & 58.8 & 61  \\
Li \textit{et al.} \cite{Li2018ECCV} & RN-101 & - & 62.9 & - \\
DeeperLab \cite{Yang2019DeeperLab} & XC-71 & 512 x 512 & \textbf{67.4} & 90 \\
FPSNet (ours) & RN-50-FPN & 512 x 512 & 57.8 & \textbf{28} \\

\end{tabular} 
\label{tab:results_main_pascal}
\end{table}

\subsection{Performance on Pascal VOC}
\begin{figure}[t]
\centering
\includegraphics[width=0.350\linewidth]{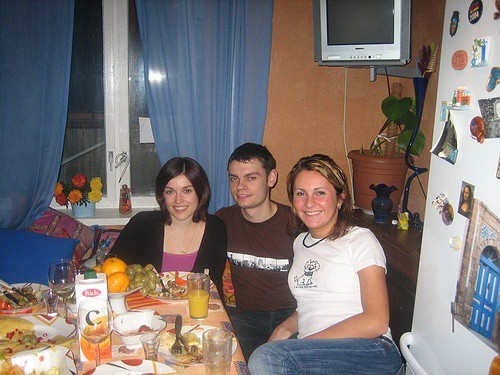}
\includegraphics[width=0.350\linewidth]{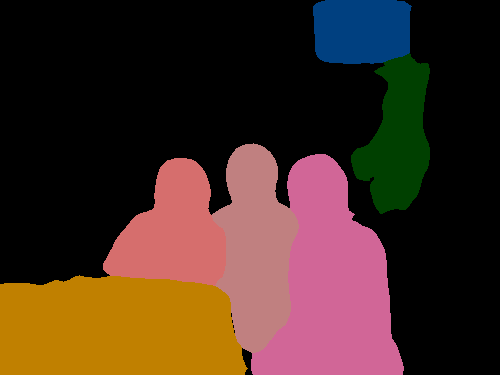}\\
\includegraphics[width=0.350\linewidth]{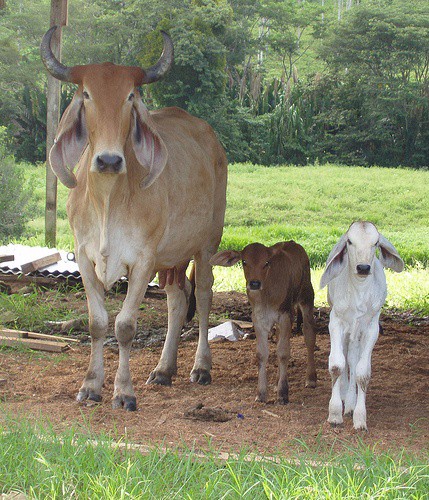}
\includegraphics[width=0.350\linewidth]{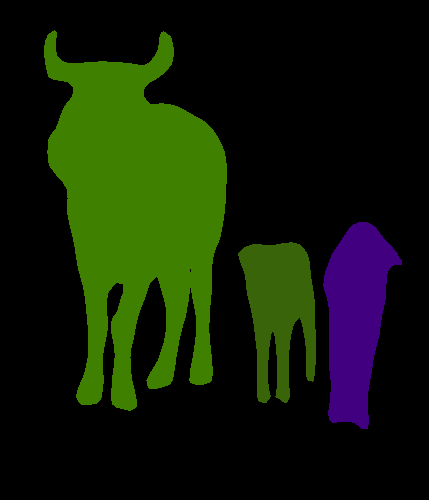}\\
\includegraphics[width=0.350\linewidth]{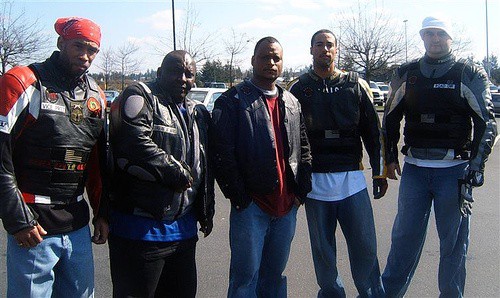}
\includegraphics[width=0.350\linewidth]{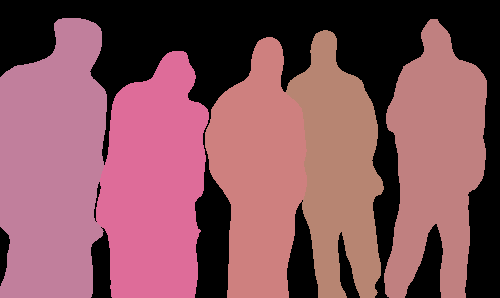}\\
\includegraphics[width=0.350\linewidth]{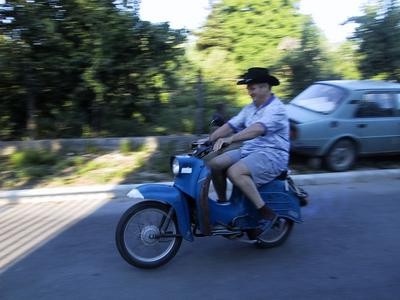}
\includegraphics[width=0.350\linewidth]{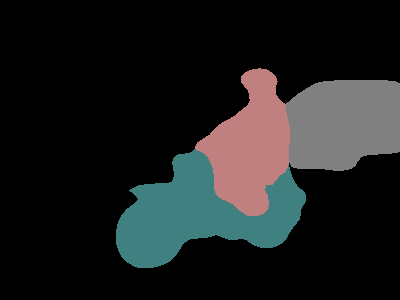}\\
\caption{Example predictions by FPSNet on the Pascal VOC 2012 validation set. Each color indicates a different \textit{things} instance.}
\label{fig:results_examples_voc}

\end{figure}

We evaluate our results on the Pascal VOC 2012, and compare with other methods in Table \ref{tab:results_main_pascal}, in terms of PQ and total prediction time. Again, it is clear that FPSNet is by far the fastest method, whilst achieving competitive PQ scores. Since this dataset only consists of \textit{things} classes, and hence the balance between \textit{stuff} and \textit{things} classes is completely different than for Cityscapes, it is possible that changing hyperparameters can substantially improve the performance. Qualitative results on the Pascal VOC 2012 validation set are shown in Figure \ref{fig:results_examples_voc}.

\section{Conclusions}
\label{sec:conclusions}
In this work, we presented FPSNet, an end-to-end framework for fast panoptic segmentation. FPSNet makes dense panoptic segmentation predictions in a fashion that does not require computationally expensive instance mask predictions or merging heuristics. This is facilitated by a novel panoptic head design and a tailored panoptic training strategy. With extensive experiments, we have shown that FPSNet is faster than existing state-of-the-art panoptic segmentation networks, and is able to achieve real-time frame rates of up to 35 fps at a resolution of 512x512 pixels. While being fast, FPSNet also achieves a competitive Panoptic Quality score of 55.1 on the Cityscapes validation set. With this work, we have made a significant step in bringing high-quality panoptic segmentation to real-time applications in robotics and intelligent vehicles.

\bibliographystyle{IEEEtran}
\bibliography{bibliography}

\begin{thebibliography}{10}
\providecommand{\url}[1]{#1}
\csname url@rmstyle\endcsname
\providecommand{\newblock}{\relax}
\providecommand{\bibinfo}[2]{#2}
\providecommand\BIBentrySTDinterwordspacing{\spaceskip=0pt\relax}
\providecommand\BIBentryALTinterwordstretchfactor{4}
\providecommand\BIBentryALTinterwordspacing{\spaceskip=\fontdimen2\font plus
\BIBentryALTinterwordstretchfactor\fontdimen3\font minus
  \fontdimen4\font\relax}
\providecommand\BIBforeignlanguage[2]{{%
\expandafter\ifx\csname l@#1\endcsname\relax
\typeout{** WARNING: IEEEtran.bst: No hyphenation pattern has been}%
\typeout{** loaded for the language `#1'. Using the pattern for}%
\typeout{** the default language instead.}%
\else
\language=\csname l@#1\endcsname
\fi
#2}}

\bibitem{Kirillov2018}
A.~Kirillov, K.~He, R.~Girshick, C.~Rother, and P.~Doll{\'a}r, ``{Panoptic
  Segmentation},'' in \emph{The IEEE Conference on Computer Vision and Pattern
  Recognition (CVPR)}, June 2019.

\bibitem{Kirillov2019}
A.~Kirillov, R.~Girshick, K.~He, and P.~Doll{\'{a}}r, ``{Panoptic Feature
  Pyramid Networks},'' in \emph{The IEEE Conference on Computer Vision and
  Pattern Recognition (CVPR)}, June 2019.

\bibitem{Xiong2019}
Y.~Xiong, R.~Liao, H.~Zhao, R.~Hu, M.~Bai, E.~Yumer, and R.~Urtasun, ``{UPSNet:
  A Unified Panoptic Segmentation Network},'' in \emph{The IEEE Conference on
  Computer Vision and Pattern Recognition (CVPR)}, June 2019.

\bibitem{Porzi2019Seamless}
L.~Porzi, S.~R. Bulo, A.~Colovic, and P.~Kontschieder, ``{Seamless Scene
  Segmentation},'' in \emph{The IEEE Conference on Computer Vision and Pattern
  Recognition (CVPR)}, June 2019.

\bibitem{Li2018TASCNet}
J.~Li, A.~Raventos, A.~Bhargava, T.~Tagawa, and A.~Gaidon, ``{Learning to Fuse
  Things and Stuff},'' \emph{arXiv preprint arXiv:1812.01192}, Dec. 2018.

\bibitem{Li2018AUNet}
Y.~Li, X.~Chen, Z.~Zhu, L.~Xie, G.~Huang, D.~Du, and X.~Wang,
  ``{Attention-Guided Unified Network for Panoptic Segmentation},'' in
  \emph{The IEEE Conference on Computer Vision and Pattern Recognition (CVPR)},
  June 2019.

\bibitem{Yang2019DeeperLab}
T.-J. Yang, M.~D. Collins, Y.~Zhu, J.-J. Hwang, T.~Liu, X.~Zhang, V.~Sze,
  G.~Papandreou, and L.-C. Chen, ``{DeeperLab: Single-Shot Image Parser},''
  \emph{{arXiv preprint arXiv:1902.05093}}, {2019}.

\bibitem{Arnab2017}
A.~{Arnab} and P.~H.~S. {Torr}, ``{Pixelwise Instance Segmentation with a
  Dynamically Instantiated Network},'' in \emph{2017 IEEE Conference on
  Computer Vision and Pattern Recognition (CVPR)}, July 2017, pp. 879--888.

\bibitem{yao2012describing}
S.~Fidler, J.~Yao, and R.~Urtasun, ``{Describing the scene as a whole: Joint
  object detection, scene classification and semantic segmentation},'' in
  \emph{2012 IEEE Conference on Computer Vision and Pattern Recognition
  (CVPR)}, June 2012, pp. 702--709.

\bibitem{tu2005image}
Z.~Tu, X.~Chen, A.~Yuille, and S.~Zhu, ``{Image parsing: Unifying segmentation,
  detection, and recognition},'' \emph{International Journal of computer
  vision}, vol.~63, no.~2, pp. 113--140, 2005.

\bibitem{Li2018ECCV}
Q.~Li, A.~Arnab, and P.~Torr, ``{Weakly- and Semi-Supervised Panoptic
  Segmentation},'' in \emph{The European Conference on Computer Vision (ECCV)},
  Sept. 2018.

\bibitem{DeGeus2018}
D.~de~Geus, P.~Meletis, and G.~Dubbelman, ``{Panoptic Segmentation with a Joint
  Semantic and Instance Segmentation Network},'' \emph{arXiv preprint
  arXiv:1809.02110}, Sept. 2018.

\bibitem{He2017}
K.~{He}, G.~{Gkioxari}, P.~{Doll{\'a}r}, and R.~{Girshick}, ``{Mask R-CNN},''
  in \emph{2017 IEEE International Conference on Computer Vision (ICCV)}, Oct.
  2017, pp. 2980--2988.

\bibitem{Zhao2017}
H.~{Zhao}, J.~{Shi}, X.~{Qi}, X.~{Wang}, and J.~{Jia}, ``{Pyramid Scene Parsing
  Network},'' in \emph{2017 IEEE Conference on Computer Vision and Pattern
  Recognition (CVPR)}, July 2017, pp. 6230--6239.

\bibitem{Liu2019}
H.~Liu, C.~Peng, C.~Yu, J.~Wang, X.~Liu, G.~Yu, and W.~Jiang, ``{An End-To-End
  Network for Panoptic Segmentation},'' in \emph{The IEEE Conference on
  Computer Vision and Pattern Recognition (CVPR)}, June 2019.

\bibitem{Yang2019}
T.~Yang, M.~Collins, Y.~Zhu, J.~Hwang, T.~Liu, X.~Zhang, V.~Sze, G.~Papandreou,
  and L.~Chen, ``{DeeperLab: Single-Shot Image Parser},'' \emph{arXiv preprint
  arXiv:1902.05093}, Feb. 2019.

\bibitem{Kirillov2017InstanceCut}
A.~Kirillov, E.~Levinkov, B.~Andres, B.~Savchynskyy, and C.~Rother,
  ``{InstanceCut: From Edges to Instances With MultiCut},'' in \emph{The IEEE
  Conference on Computer Vision and Pattern Recognition (CVPR)}, July 2017.

\bibitem{uhrig2016pixel}
J.~Uhrig, M.~Cordts, U.~Franke, and T.~Brox, ``Pixel-level encoding and depth
  layering for instance-level semantic labeling,'' in \emph{German Conference
  on Pattern Recognition}.\hskip 1em plus 0.5em minus 0.4em\relax Springer,
  2016, pp. 14--25.

\bibitem{Nair2010ReLU}
V.~Nair and G.~Hinton, ``{Rectified Linear Units Improve Restricted Boltzmann
  Machines},'' in \emph{Proceedings of the 27th International Conference on
  Machine Learning}, 2010, pp. 807--814.

\bibitem{Ioffe2015BatchNorm}
S.~Ioffe and C.~Szegedy, ``{Batch Normalization: Accelerating Deep Network
  Training by Reducing Internal Covariate Shift},'' in \emph{Proceedings of the
  32nd International Conference on Machine Learning}, 2015, pp. 448--456.

\bibitem{Lin2017RetinaNet}
T.~{Lin}, P.~{Goyal}, R.~{Girshick}, K.~{He}, and P.~{Doll{\'a}r}, ``{Focal
  Loss for Dense Object Detection},'' in \emph{2017 IEEE International
  Conference on Computer Vision (ICCV)}, Oct. 2017, pp. 2999--3007.

\bibitem{He2015}
K.~He, X.~Zhang, S.~Ren, and J.~Sun, ``{Deep Residual Learning for Image
  Recognition},'' in \emph{2016 IEEE Conference on Computer Vision and Pattern
  Recognition (CVPR)}, June 2016, pp. 770--778.

\bibitem{Lin2017}
T.~{Lin}, P.~{Doll{\'a}r}, R.~{Girshick}, K.~{He}, B.~{Hariharan}, and
  S.~{Belongie}, ``{Feature Pyramid Networks for Object Detection},'' in
  \emph{2017 IEEE Conference on Computer Vision and Pattern Recognition
  (CVPR)}, July 2017, pp. 936--944.

\bibitem{DeGeus2019}
D.~{de Geus}, P.~{Meletis}, and G.~{Dubbelman}, ``{Single Network Panoptic
  Segmentation for Street Scene Understanding},'' in \emph{2019 IEEE
  Intelligent Vehicles Symposium (IV)}, June 2019, pp. 709--715.

\bibitem{Cordts2016Cityscapes}
M.~Cordts, M.~Omran, S.~Ramos, T.~Rehfeld, M.~Enzweiler, R.~Benenson,
  U.~Franke, S.~Roth, and B.~Schiele, ``{The Cityscapes Dataset for Semantic
  Urban Scene Understanding},'' in \emph{2016 IEEE Conference on Computer
  Vision and Pattern Recognition (CVPR)}, June 2016, pp. 3213--3223.

\bibitem{Everingham2010Pascal}
M.~Everingham, L.~Van~Gool, C.~K.~I. Williams, J.~Winn, and A.~Zisserman,
  ``{The Pascal Visual Object Classes (VOC) Challenge},'' \emph{International
  Journal of Computer Vision}, vol.~88, no.~2, pp. 303--338, June 2010.

\bibitem{Bharath2011SBD}
B.~Hariharan, P.~Arbelaez, L.~Bourdev, S.~Maji, and J.~Malik, ``{Semantic
  Contours from Inverse Detectors},'' in \emph{International Conference on
  Computer Vision (ICCV)}, 2011.

\bibitem{Chen2018}
L.~Chen, G.~Papandreou, I.~Kokkinos, K.~Murphy, and A.~L. Yuille, ``{DeepLab:
  Semantic Image Segmentation with Deep Convolutional Nets, Atrous Convolution,
  and Fully Connected CRFs},'' \emph{IEEE Transactions on Pattern Analysis and
  Machine Intelligence}, vol.~40, no.~4, pp. 834--848, Apr. 2018.

\bibitem{Deng2009ImageNet}
J.~Deng, W.~Dong, R.~Socher, L.~Li, K.~Li, and L.~Fei-Fei, ``{ImageNet: A
  large-scale hierarchical image database},'' in \emph{2009 IEEE Conference on
  Computer Vision and Pattern Recognition}, June 2009, pp. 248--255.

\end{thebibliography}

\addtolength{\textheight}{-12cm}   %

\end{document}